\documentclass[conference]{IEEEtran}
\IEEEoverridecommandlockouts
\usepackage{cite}
\usepackage{amsmath,amssymb,amsfonts}
\usepackage{algorithmic}
\usepackage{graphicx}
\usepackage{textcomp}
\usepackage{xcolor}
\def\BibTeX{{\rm B\kern-.05em{\sc i\kern-.025em b}\kern-.08em
    T\kern-.1667em\lower.7ex\hbox{E}\kern-.125emX}}

\usepackage[hidelinks]{hyperref}
\usepackage{amsmath}
\usepackage{amssymb} %
\usepackage[caption=false]{subfig}
\usepackage[capitalise]{cleveref}
\usepackage{csquotes}   %
\usepackage[final]{listings}
\usepackage{pifont} %
\newcommand{\cmark}{\ding{51}}%
\newcommand{\xmark}{\ding{55}}%

\usepackage{flushend}
\usepackage{enumitem}

\setlist[itemize]{noitemsep, topsep=0pt}
\usepackage{booktabs}
\let\originalbottomrule\bottomrule
\renewcommand{\bottomrule}{\addlinespace[0pt]\originalbottomrule}
\let\originalmidrule\midrule
\renewcommand{\midrule}{\addlinespace[0pt]\originalmidrule}
\usepackage{multirow}
\usepackage{tabularx}   %

\usepackage{textcomp}

\begin{document}

\title{Tuning the Tuner: Introducing Hyperparameter Optimization for Auto-Tuning
\thanks{The CORTEX project has received funding from the Dutch Research Council (NWO) in the framework of the NWA-ORC Call (file \#NWA.1160.18.316).
The ESiWACE3 project has received funding from the European High Performance Computing Joint Undertaking (JU) under grant agreement No 101093054. We acknowledge the EuroHPC Joint Undertaking for awarding this project access to the EuroHPC supercomputer LUMI, hosted by CSC (Finland) and the LUMI consortium through a EuroHPC Regular Access call.}
}

\author{
    \IEEEauthorblockN{Floris-Jan Willemsen}
    \IEEEauthorblockA{
        \textit{LIACS, Leiden University} \\
        \textit{Netherlands eScience Center} \\
        Leiden \& Amsterdam, the Netherlands \\
        ORCID: 0000-0003-2295-8263
    }
\and
    \IEEEauthorblockN{Rob V. van Nieuwpoort}
    \IEEEauthorblockA{
        \textit{LIACS, Leiden University} \\
        Leiden, the Netherlands \\
        ORCID: 0000-0002-2947-9444
    }
\and
    \IEEEauthorblockN{Ben van Werkhoven}
    \IEEEauthorblockA{
        \textit{LIACS, Leiden University} \\
        \textit{Netherlands eScience Center} \\
        Leiden \& Amsterdam, the Netherlands \\
        ORCID: 0000-0002-7508-3272
    }
}

\maketitle
\thispagestyle{plain}
\pagestyle{plain}

\begin{abstract}
Automatic performance tuning (auto-tuning) is widely used to optimize performance-critical applications across many scientific domains by finding the best program variant among many choices. 
Efficient optimization algorithms are crucial for navigating the vast and complex search spaces in auto-tuning. 
As is well known in the context of machine learning and similar fields, hyperparameters critically shape optimization algorithm efficiency. 
Yet for auto-tuning frameworks, these hyperparameters are almost never tuned, and their potential performance impact has not been studied.

We present a novel method for general hyperparameter tuning of optimization algorithms for auto-tuning, thus "tuning the tuner".
In particular, we propose a robust statistical method for evaluating hyperparameter performance across search spaces, publish a FAIR data set and software for reproducibility, and present a simulation mode that replays previously recorded tuning data, lowering the costs of hyperparameter tuning by two orders of magnitude.
We show that even limited hyperparameter tuning can improve auto-tuner performance by 94.8\% on average, and establish that the hyperparameters themselves can be optimized efficiently with meta-strategies (with an average improvement of 204.7\%), demonstrating the often overlooked hyperparameter tuning as a powerful technique for advancing auto-tuning research and practice.

\end{abstract}


    \section{Introduction}
\label{sec:introduction}

\IEEEPARstart{A}{utomatic} performance tuning, or auto-tuning, is a widely established method for optimizing the performance of applications in many scientific domains, including 
radio astronomy~\cite{price2016optimizing, sclocco2020amber, schoonhoven2022going, oostrum2025tcbf},
image processing~\cite{martinez2011automatic, wang2016auto, falch2017machine},
fluid dynamics~\cite{dong2014step, chen2015angel, rojek2019machine}, and
climate modeling~\cite{nan2014cesmtuner, zhang2018automatic, heldens2023kernel}.
Auto-tuning automates the process of exploring the myriad of implementation choices that arise in performance optimization, such as the number of threads, tile sizes used in loop blocking, and other code optimization parameters~\cite{balaprakash2017autotuning}.
At the heart of the auto-tuning method is a {\em search space} of functionally-equivalent {\em code variants} that is explored by an {\em optimization algorithm}.

Together, these code variants constitute vast search spaces that are infeasible to search by hand~\cite{pruning,scloccoAutoTuningDedispersionManyCore2014,lessonsLearnedGPU2020} and would have to be searched over and over again as the application is executed on different hardware or different input data sets and sizes~\cite{CLBlast2018,vanWerkhoven2014optimizing,lawson2019cross,KTTBenchmark}.
Therefore, a key problem in auto-tuning research is the development of optimization algorithms that can efficiently navigate the auto-tuning search space~\cite{vanwerkhovenKernelTunerSearchoptimizing2019, schoonhovenBenchmarkingOptimizationAlgorithms2022}.

The effectiveness of optimization algorithms is strongly influenced by their {\em hyperparameters}, i.e., settings that control how the algorithm explores the search space.
Hyperparameter tuning is essential in fields such as machine learning to determine optimal algorithm settings~\cite{bardenet2013collaborative}, e.g., learning rates~\cite{joy2016hyperparameter}, neural architectures~\cite{li2020system}, and regularization parameters~\cite{bergstra2011algorithms}.
While hyperparameter tuning has been successfully applied in a wide variety of areas, it has to the best of our knowledge not been applied to optimization algorithms for auto-tuning.
For example, the surveys on auto-tuning in High-Performance Computing by Balaprakash et al.~\cite{balaprakash2017autotuning}, as well as the comprehensive survey by Ashouri et al.~\cite{ashouri2018survey} on machine learning in auto-tuning do not mention hyperparameter tuning at all.
As such, the potential impact of hyperparameter tuning on the performance of optimization algorithms for auto-tuning has not been quantified or studied, and no generic auto-tuning framework currently supports it.

To bridge this gap and enable hyperparameter tuning for the field of auto-tuning, we need to overcome two key challenges:
\begin{enumerate}[leftmargin=*,topsep=0pt,itemsep=0pt,partopsep=0pt, parsep=0pt]
\item The hyperparameter tuner must optimize performance across different auto-tuning search spaces to achieve good generalization across different applications and hardware architectures. However, quantifying overall performance across different applications and architectures using different performance metrics is nontrivial.
\item To achieve robust general performance estimates of stochastic optimization algorithms currently necessitates a high number of repeated tuning runs, which makes hyperparameter tuning prohibitively expensive.
\end{enumerate}

\noindent In this work, we address these challenges and introduce a novel approach to hyperparameter tuning tailored to the problem of auto-tuning that we refer to as "tuning the tuner". 
We make the following contributions:
\begin{itemize}[leftmargin=*,topsep=0pt,itemsep=0pt,partopsep=0pt, parsep=0pt]
    \item We propose a systematic and general method for hyperparameter tuning of optimization algorithms for auto-tuning. %
    \item We propose a novel {\em simulation mode} that accelerates hyperparameter tuning by simulating optimization runs, yielding a $\sim130\times$ speedup.
    \item We provide a FAIR dataset of general auto-tuning data that is well-balanced in applications and hardware, lowering barriers to entry and reducing energy and resource costs.
    \item We present the first evaluation of the impact of hyperparameter tuning on the performance of optimization algorithms for auto-tuning, with an average performance improvement of 94.8\% on a limited, exhaustively evaluated set of hyperparameters.
    \item We present the first results on the effectiveness of meta-strategies for efficient hyperparameter tuning for auto-tuning, enabling application beyond exhaustive tuning, with an average improvement of 204.7\% on an extended set of hyperparameters tuned with a meta-strategy. 
\end{itemize}
All contributions are implemented in the open-source Kernel Tuner~\cite{vanwerkhovenKernelTunerSearchoptimizing2019} and Autotuning Methodology frameworks~\cite{methodologyPaper}.

Given the possible confusion in terminology in distinguishing between auto-tuning and hyperparameter tuning, we will use \textit{kernel configuration} to refer to the specific settings of a to-be-tuned computational kernel, and \textit{hyperparameter configuration} for the specific settings of an optimization algorithm. 
The rest of this work is structured as follows. \cref{sec:related_work} discusses related work, \cref{sec:impl} describes our method and implementation, \cref{sec:evaluation} evaluates the diverse aspects of our method, and \cref{sec:conclusion_futurework} concludes the paper.

    \section{Related Work}

\label{sec:related_work}
\begin{table*}[tb]
\small
\centering
\caption{
Overview of auto-tuning frameworks, their open source status, supported optimization algorithms, and support for configuring hyperparameters without modifying source code. SA = Simulated Annealing. PSO = Particle Swarm Optimization. 
}
\label{tab:autotuners}
\setlength\tabcolsep{4pt}
\begin{tabularx}{\textwidth}{l|c|c|X|c}
\toprule
Framework	& Open Source	& Active	& Optimization algorithms	& Hyperparameters	\\
\midrule

OpenTuner~\cite{OpenTuner}	& \cmark	& \xmark	& SA, PSO, Multi-armed Bandit, Evolutionary methods, Local Search & \xmark	\\ \hline
KTT~\cite{KTT}	& \cmark	& \cmark	& Markov Chain Monte Carlo, Profile-based search	& \xmark	\\ \hline
ATF~\cite{ATF} / PyATF~\cite{pyATF} & \cmark	& \cmark	& SA, Differential Evolution, Multi-armed Bandit, Local Search & \xmark	\\ \hline
GPTune~\cite{liuGPTuneMultitaskLearning2021} & \cmark & \cmark	& Bayesian Optimization & \xmark \\ \hline
AUMA~\cite{AUMA}	& \cmark	& \xmark	& K-Bagging Neural Networks	& \cmark (File-based)		\\ \hline
BaCO~\cite{BaCO2024} & \cmark & \xmark	& Bayesian Optimization, Exhaustive	& \cmark (File-based)	\\ \hline
ytopt~\cite{ytopt} & \cmark & \cmark & Bayesian Optimization & \cmark (command line)	\\ \hline
CLTune~\cite{CLTune}	& \cmark	& \xmark	& SA, PSO, Neural Network & \cmark (API-based)	\\ \hline
Kernel Tuner~\cite{vanwerkhovenKernelTunerSearchoptimizing2019}	& \cmark & \cmark   & 20+ global and local optimization algorithms, including Annealing, Evolutionary, and Swarm-based methods, and Bayesian Optimization	& \cmark (API-based)	\\
\bottomrule
\end{tabularx}
\end{table*}

There are many different automated approaches to improving the performance of software that are collectively referred to as auto-tuning. For a survey of different uses of auto-tuning in high-performance computing, see Balaprakash et al.~\cite{balaprakash2017autotuning}. 
As stated in the previous section, to the best of our knowledge, the application and impact of hyperparameter tuning in the context of auto-tuning is largely unstudied.
As such, this section reviews the support for optimization algorithms and the possibility of controlling their hyperparameters in the current state-of-the-art generic auto-tuners, and reviews a handful of studies that include a preliminary evaluation of hyperparameters in auto-tuning.

\Cref{tab:autotuners} presents an overview of generic auto-tuning frameworks, the optimization algorithms they implement, and their support for setting hyperparameters. As we can see, many of these frameworks implement advanced search strategies, but few offer a practical interface to configure hyperparameters without modifying the tuner itself. For example,
OpenTuner~\cite{OpenTuner}, %
KTT~\cite{KTT}, %
ATF~\cite{ATF}, PyATF~\cite{pyATF}, %
and
GPTune~\cite{liuGPTuneMultitaskLearning2021} %
do not support specifying the hyperparameters without source code modifications.

Falch and Elster have presented the AUMA~\cite{AUMA} %
auto-tuner to improve the performance portability of OpenCL applications across various 
parallel architectures, including Intel CPUs and GPUs, as well as AMD and Nvidia GPUs. 
AUMA uses a machine-learning approach to predict the subset of the search space that is likely to contain high-performant code variants.
Their method implements one hyperparameter called the {\em threshold} that controls when to stop tuning. They evaluate their methods for four different thresholds, showing that the optimal value strongly depends on the tuned application. %

The BaCO~\cite{BaCO2024} %
auto-tuner supports setting the hyperparameters in a JSON file as part of the specification of a tuning problem. However, BaCO is not evaluated or tested with different hyperparameters in the accompanying paper~\cite{BaCO2024}.

Nugteren and Codreanu~\cite{CLTune} implemented Simulated Annealing (SA), Particle Swarm Optimization (PSO), as well as a neural network-based optimizer, in CLTune. CLTune provides an API-based interface for controlling the hyperparameters, and the authors do a very limited evaluation of hyperparameter sensitivity. They evaluate three different values for the {\em temperature} hyperparameter of SA, as well as two different values for the {\em swarm size} of PSO for auto-tuning a 2D convolution kernel on four GPUs.
However, their evaluation shows only minimal performance differences, with no optimization algorithm implemented in CLTune outperforming random search. 
The authors do note that "the evaluated search strategies are based on stochastic variables, we cannot draw conclusions from this limited set of experiments", which underlines the need for robust hyperparameter tuning methods.

Kernel Tuner~\cite{vanwerkhovenKernelTunerSearchoptimizing2019} %
is the only actively developed open-source auto-tuner with an API for controlling the hyperparameters. As such, we have chosen Kernel Tuner as the tuner in which we implement our method.

The most closely related work is perhaps the survey of optimization algorithms for auto-tuning by Schoonhoven et al.~\cite{schoonhovenBenchmarkingOptimizationAlgorithms2022}, in which a limited hyperparameter tuning is conducted to compare optimization algorithms for auto-tuning.
They measure performance by counting head-to-head wins between two algorithms on individual problems and select hyperparameter tuning configurations by repeatedly combining configurations with the most wins from individual problems.
However, generalization is not addressed, and the performance impact of hyperparameter tuning is not evaluated in their study.

\section{Design and Implementation}\label{sec:impl}

In this work, we present a method for general hyperparameter tuning for auto-tuning; this section provides an overview of its design and implementation.
\Cref{subsec:impl_context} provides an introduction of the auto-tuning problem, leading into \cref{subsec:impl_hyperparameter_tuning} which introduces the proposed hyperparameter tuning method.
\Cref{subsec:impl_feasibility} presents our simulation mode to make hyperparameter tuning feasible. %
\Cref{subsec:impl_fair_data} presents our FAIR data set for benchmarking auto-tuner performance, and \cref{subsec:impl_kernel_tuner} discusses the implementation.

\subsection{The auto-tuning problem} \label{subsec:impl_context}

Auto-tuning involves optimizing an application or {\em kernel} $K_i$ on a target system $G_j$ for input data $I_k$ to maximize performance $f_{G_j,I_k}(K_i)$. 
The auto-tuner constructs a search space $\mathcal{X}$ by considering all tunable parameters and their valid values, subject to user-defined constraints~\cite{willemsen2025SearchSpaceConstruction}. 
The objective is to determine the optimal configuration, or more formally (assuming minimization) as follows: 
\begin{equation} \label{eq:autotuning}
x^\star = \underset{x\in\mathcal{X}}{\text{arg min}} \, f_{G_j,I_k}(K_{i,x}).
\end{equation}

The evaluation of each configuration takes time, as it needs to be compiled and executed on the target system. 
Search spaces in auto-tuning are often large, discontinuous, non-convex, and irregular. 
These characteristics make them infeasible to search by hand and demand carefully chosen optimization algorithms. 
If the algorithm is not well adapted to the search space, the number of required evaluations becomes excessive, limiting the practical usability of the auto-tuner. 

\subsection{Generalized Hyperparameter Tuning} \label{subsec:impl_hyperparameter_tuning}

Hyperparameter tuning adjusts the settings of an optimization algorithm to enhance its efficiency across a domain of problems. 
Examples of hyperparameters include the population size in evolutionary algorithms, neighbor selection in local search methods, and temperature schedules in annealing-based approaches. 
While auto-tuners typically optimize performance-related parameters for a specific application on a specific target system, our hyperparameter tuning approach aims for \emph{general optimized performance across multiple tuning problems}.

In earlier work, we presented a community-driven methodology for evaluating optimization algorithms in auto-tuning~\cite{methodologyPaper}, which we extend to hyperparameter tuning. 
The methodology provides a systematic approach to comparing optimization algorithms across auto-tuning search spaces. 
It defines a performance score that quantifies an optimization algorithm's performance over the passed time relative to a calculated baseline, typically random search, to have consistent, objective-independent, transparent, and comparable behavior across search spaces. 

On an individual search space, this approach first defines a budget and performance baseline adapted to the search space characteristics. 
The optimization algorithms to be compared are then executed multiple times to account for noise, after which the difference between the baseline and the average best performance of each optimization algorithm is compared at fixed, equidistant time intervals relative to the budget. 
The result is a smooth performance curve over time, %
instead of relying solely on final performance. %
By adapting the budget and baseline to reliable search space characteristics, we can compare and aggregate these performance curves across search spaces. 
We will thus use the mean of these aggregated performance curves as a performance score $\mathcal{P}$ for general hyperparameter tuning of an optimization algorithm, where a higher score is better overall performance. 

More formally, for a given time sampling point $t$ the performance $\mathcal{P}_t$ of an optimization algorithm $\mathcal{F}$ can be computed:
\vspace{-0.2cm}
\begin{equation} \label{eq:performance_curve}
\mathcal{P}(\mathcal{F},G_j,K_i,I_k)_t = \frac{\mathcal{S}_{\text{baseline}}(t) - \mathcal{F}(G_j,K_i,I_k)_t}{S_{\text{baseline}}(t) - \mathcal{S}_{\text{opt}}}
\end{equation}
where $\mathcal{S}(G_j,K_i,I_k)$ are the search space characteristics given a target system $G_j$, kernel $K_i$, and input data $I_k$. 
Here, $\mathcal{F}_t$ is the best objective value found so far by the optimization algorithm, $\mathcal{S}_{\text{baseline}}(t)$ is the performance of the baseline method, and $\mathcal{S}_{\text{opt}}$ is the known optimal value in the search space. 
This yields $\mathcal{P}_t = 0$ when performance equals the baseline and $\mathcal{P}_t = 1$ when the optimum has been found. 

To avoid distorting the metric with large variations in search space difficulty, evaluations are limited to the time at which the baseline reaches a set cutoff percentile between the median and the optimum - typically somewhere around 95\% - referred to as the budget. 
With this method, the performance curves are relative to the same baseline and optimum, the cutoff provides a consistent reference point, and the sampling points are equidistant. 
The performance curves can thus be meaningfully aggregated from different search spaces by taking the mean score of all curves at each $t$, resulting in the aggregate performance curve over all search spaces for the optimization algorithm. 
The aggregate performance score is then obtained by averaging over the set of discrete time sampling points $\mathcal{T}$, or more formally:
\vspace{-0.4cm}

\begin{equation} \label{eq:performance_score}
\mathcal{P}(\mathcal{F},K,G,I) = \frac{1}{|\mathcal{T}|} \sum_{t \in \mathcal{T}} \frac{\displaystyle \sum_{K_i \in K} \sum_{G_j \in G} \sum_{I_k \in I} \mathcal{P}(\mathcal{F}, G_j, K_i, I_k)_t}{|K||G||I|}
\end{equation}
The kernels $K$, target systems $G$, and inputs $I$ are the collections of $K_i$, $G_j$, and $I_k$ of \cref{eq:autotuning} on which the hyperparameter tuning is conducted, also referred to as the training set. 
This aggregate score enables robust comparison of optimization algorithms by capturing both the quality of the configurations found as well as the time taken to do so. 
As such, a difference when comparing two performance scores can indicate a difference in the quality of configurations found, the time taken to do so, or a combination of both. 

The hyperparameter tuning problem can thus be formalized:
\begin{equation} \label{eq:hyperparamtuning}
{h^\star} = \underset{h\in H}{\text{arg max}} \, \mathcal{P}(\mathcal{F}_h, K, G, I)
\end{equation}
where $H$ represents the hyperparameter space, and $\mathcal{F}_h$ is the optimization algorithm configured with hyperparameters $h$.

\subsection{Hyperparameter Tuning Feasibility} \label{subsec:impl_feasibility}
It must be noted that evaluating \cref{eq:hyperparamtuning} requires that for each hyperparameter configuration $h$ in $H$, an auto-tuning experiment is run on each combination of the involved kernels, target systems, and input datasets as seen in \cref{eq:performance_score}, leading to a combinatorial explosion in computational cost. 
In addition, this requires that all systems used are available for the full duration of the hyperparameter tuning. 
Moreover, many optimization algorithms are stochastic, requiring repeated evaluations to ensure a reliable outcome of the hyperparameter tuning. 
These factors combined make executing such a hyperparameter tuning prohibitively expensive and time-consuming. 

To address this challenge, we introduce a \textit{simulation mode}. 
Instead of running the recurring auto-tuning runs required for hyperparameter tuning directly on the hardware, this simulation mode retrieves pre-collected performance data from a cache file, mimicking the behavior of real auto-tuning runs. 
For such a simulation mode to work, all configurations in a search space must have been evaluated on the actual hardware (an exhaustive or \textit{brute-force} auto-tuning search). 
This might seem counterintuitive, as we are optimizing algorithms to prevent such brute-force searches when auto-tuning in the first place. 
To illustrate why this is sensible in this case, imagine tuning the hyperparameters of a stochastic algorithm with 100 possible hyperparameter configurations for 1000 function evaluations, each run of the algorithm repeated 25 times due to stochasticity, on 3 kernels across 3 target systems. 
In addition, assume each of the 9 resulting search spaces has one million valid configurations, that each take one second to compile and execute. 
Brute-forcing each search space would take $\sim11.6$ days, or $\sim104$ compute days for all 9 search spaces, although the individual search spaces can be brute-forced in parallel.
In contrast, hyperparameter tuning a single optimization algorithm in the same scenario with live auto-tuning runs would take 260 compute days. %
And that is just for a single optimization algorithm; as seen in \cref{tab:autotuners}, most auto-tuning frameworks provide multiple optimization algorithms. 

The simulation mode approach provides several important advantages. 
First, it substantially improves efficiency: after the high one-off cost of brute-forcing, the threshold of comparison to and hyperparameter tuning of additional optimization algorithms is substantially lower; due to the discrete nature of the search spaces combined with how the budget works (\cref{subsec:impl_hyperparameter_tuning}) and multiple repeats required for stochastic optimization algorithms, configurations are likely to be revisited. %
In addition, it mitigates measurement noise: the simulation mode ensures every auto-tuning execution is reproducible, preventing additional noise in the process.
Finally, it reduces energy use and resource contention: once the full search space is evaluated, hyperparameter tuning can be performed without costly compilation and execution on the target system, and without occupying those resources during the hyperparameter tuning. 

To ensure the simulation mode is representative of real-world auto-tuning performance, we collect execution data across a diverse set of tuning problems and hardware. 
This data serves as a foundation for evaluating hyperparameter tuning strategies without the need for real-time benchmarking, enabling rapid experimentation and algorithm comparison and reducing cost and energy consumption.
In addition, making this publicly available provides and promotes a reproducible record that can be used and contributed to by the community.

\subsection{FAIR benchmark data for Hyperparameter Tuning}
\label{subsec:impl_fair_data}

To enable reproducible and scalable hyperparameter tuning of optimization algorithms for GPU auto-tuning in particular, we have developed and curated a dataset of fully brute-forced search spaces across a diverse set of kernels and GPU architectures. 
This dataset, referred to as the \textit{Benchmark Hub for Auto-Tuning}, serves as a community resource to facilitate benchmarking and collaborative research. 
It consists of 24 exhaustively evaluated search spaces, formed by the Cartesian product of four real-world GPU kernels across six GPUs. 
Each kernel configuration in these search spaces has been executed 32 times to mitigate measurement noise, with both the average and raw values present in the data.

The four applications are the \textit{dedispersion}, \textit{convolution}, \textit{hotspot}, and \textit{GEMM} kernels as described by~\cite{lurati2024bringing}, widely used in astronomy, image processing, material science, and linear algebra, respectively.
Dedispersion is a signal-processing kernel that reconstructs radio signals distorted by interstellar dispersion by applying a range of dispersion measures to time-domain samples across multiple frequency channels. 
The 2D convolution stencil kernel performs image filtering by computing weighted sums over image regions. %
Hotspot is a thermal simulation stencil kernel that estimates processor temperature by iteratively solving differential equations based on simulated power and initial temperature inputs, producing a temperature grid as output. %
GEMM (General Matrix-Matrix Multiplication) is a linear algebra operation implemented in CLBlast that computes $C = \alpha A \cdot B + \beta C$ for large dense matrices. 
These applications also bring diversity in their performance characteristics; e.g., dedispersion and hotspot are generally bandwidth-bound, convolution and GEMM are generally compute-bound. 

The six GPUs are an AMD MI250X of the LUMI supercomputer, and an AMD W6600, AMD W7800, Nvidia A6000, Nvidia A4000, and Nvidia A100 of the DAS-6 supercomputer~\cite{DASMediumScaleDistributedSystem}. 
The time taken to execute the exhaustive brute-force search for each search space is listed in \cref{tab:bruteforce-times}, for a total of just over 962 hours to compute the entire dataset. 

\begin{table}[tb]
\centering
\caption{Brute-force execution times in hours for each search space.}
\label{tab:bruteforce-times}
\begin{tabularx}{\linewidth}{l|X|X|X|X|X|X}
\toprule
 Application   &   A100 &   A4000 &   A6000 &   MI250X &   W6600 &   W7800 \\
\midrule
 Dedispersion  &   10.2 &    19.9 &    12.2 &     19   &    21.4 &    10.4 \\
 Convolution   &    3.4 &     3.5 &     3.5 &      2.9 &     4.5 &     2   \\
 Hotspot       &  100.9 &    32.9 &    25.5 &     51   &    73.1 &    50.3 \\
 GEMM          &   44.7 &    58.4 &    40   &     61.2 &   250.6 &    60.9 \\
\bottomrule
\end{tabularx}
\end{table}

In alignment with the vision outlined in \textit{FAIR sharing of data in autotuning research}~\cite{autotuningFAIRT1T4}, our dataset adheres to the FAIR principles:
\begin{itemize}
    \item \textbf{Findable:} All benchmark data and associated scripts are publicly available through Zenodo\footnote{\url{https://doi.org/10.5281/zenodo.15364124}} and a dedicated GitHub repository\footnote{\url{https://github.com/AutoTuningAssociation/benchmark_hub}}, with clear structure, documentation, and Digital Object Identifiers (DOIs) for unique versions.
    \item \textbf{Accessible:} The data is open access, licensed for reuse, and includes versioned kernel code, configuration files, and benchmark results.
    \item \textbf{Interoperable:} Input data is stored in the \textit{T1} format, while output data follows the \textit{T4} format, as stipulated in~\cite{autotuningFAIRT1T4}. 
    These standardized JSON-based formats are compatible with various auto-tuning frameworks such as KTT~\cite{KTT}, Kernel Tuner~\cite{vanwerkhovenKernelTunerSearchoptimizing2019}, as well as the Autotuning Methodology software package~\cite{methodologyPaper}.
    \item \textbf{Reusable:} The dataset is directly usable in simulation mode, enabling evaluation of optimization strategies and hyperparameter tuning \emph{without requiring access to the original target systems}. Integration and utility scripts are provided to facilitate reuse and extension.
\end{itemize}

The repository is structured to support contributions: new kernels or brute-forced search spaces on additional GPUs can be submitted by the community. 
Each kernel directory includes the original source code, T1 input description, and tuning script. 
Corresponding output data is provided per GPU in both the original tuner's format and the interoperable T4 format. 
To optimize storage and portability, output files are compressed and decompressed automatically. %

This FAIR-compliant dataset empowers scalable experimentation, enables generalization across a diverse range of applications and GPUs, reduces resource usage, encourages collaboration, and lowers the barrier to entry for reproducible research in auto-tuning.

\subsection{Integration with Kernel Tuner} \label{subsec:impl_kernel_tuner}
To demonstrate the real-world impact of our proposed method, we implement it in an actively developed open-source auto-tuning framework. 
Kernel Tuner is a Python-based auto-tuner designed to optimize GPU kernels for various objectives, such as execution time and energy efficiency~\cite{vanwerkhovenKernelTunerSearchoptimizing2019}. 
It supports CUDA, HIP~\cite{lurati2024bringing}, OpenCL, and OpenACC for both C and Fortran, making it a versatile tool for GPU developers. 
With over 20 optimization algorithms, Kernel Tuner provides flexibility in searching large optimization spaces efficiently~\cite{schoonhovenBenchmarkingOptimizationAlgorithms2022}. 
The wide variety of implemented optimization algorithms and provided interface for setting hyperparameter configurations make Kernel Tuner an appropriate candidate for implementing our hyperparameter tuning method. 

\begin{figure}[htb]
    \centering
    \includegraphics[width=0.95\linewidth]{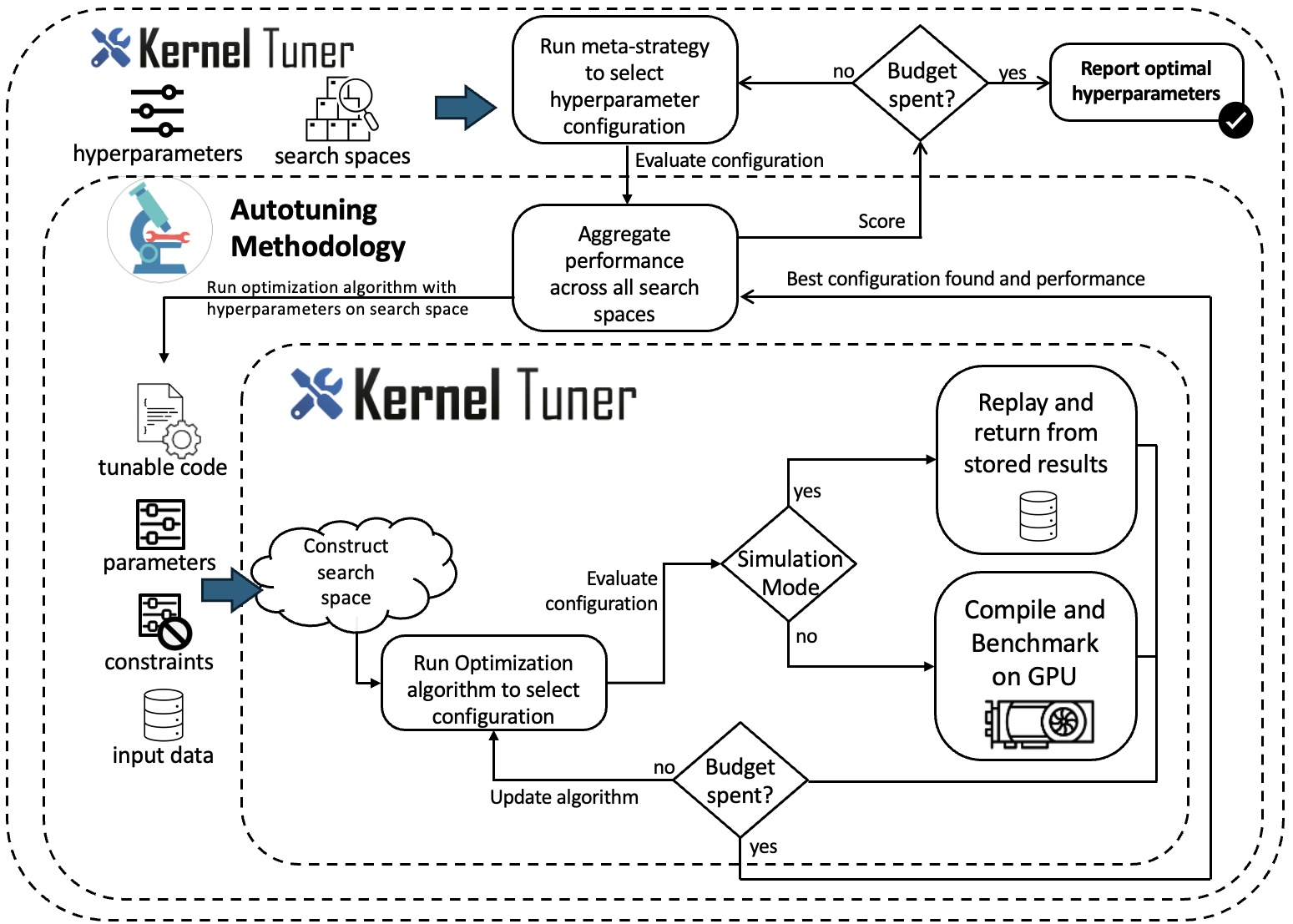}
    \caption{The hyperparameter tuning pipeline for auto-tuning. Kernel Tuner's hyperparameter tuning functionality (the outermost layer) calls the autotuning methodology software to get an aggregate performance score, which is obtained by running the optimization algorithm on various search spaces.}
    \label{fig:hyperparametertuning-pipeline}
\end{figure}

An abstract visualization of our hyperparameter tuning pipeline is shown in \cref{fig:hyperparametertuning-pipeline}. 
The hyperparameter tuning we implemented in Kernel Tuner uses the autotuning methodology~\cite{methodologyPaper} software package to request the performance score $\mathcal{P}$ of \cref{eq:performance_score} for an optimization algorithm with a selected hyperparameter configuration across the search spaces selected for training. 
The autotuning methodology package in turn uses our new simulation mode to run the optimization algorithm with the hyperparameter configuration on each of the search spaces. 
This process is repeated until the set tuning budget is spent or all hyperparameter configurations have been evaluated. 

We implemented the simulation mode by extending Kernel Tuner’s existing checkpointing mechanism. 
Each segment in the process of evaluating an auto-tuning configuration is registered, such as the time spent by the optimization algorithm, compilation, execution, and framework overhead, providing a trace of an auto-tuning run that can be replayed.  
For each auto-tuning configuration evaluated, 
\cref{fig:hyperparametertuning-pipeline} illustrates how the simulation mode integrates with Kernel Tuner’s architecture.
A dedicated simulation runner serves cached performance data, ensuring that the simulated execution accurately reflects real-world tuning times. 
Additionally, we modify the stopping criteria of optimization algorithms to account for simulated time, ensuring fair comparisons across different tuning strategies.
During simulation mode, the optimization algorithm selects a configuration, upon which the trace recorded earlier is replayed to get the result and update the tuning budget as if it had been executed. 
From the point of view of the optimization algorithm, there is no perceivable difference between live tuning and the simulation mode. 
The simulation mode is integrated into Kernel Tuner’s workflow, benefiting users by allowing switching between live and simulated tuning runs. 
By enabling efficient and repeatable evaluation of optimization algorithms, the simulation mode makes large-scale hyperparameter tuning feasible without excessive resource consumption.

To efficiently explore hyperparameter configurations, we also extended Kernel Tuner to support using its optimization algorithms as meta-strategies for hyperparameter tuning. 
This enables more informed searches for hyperparameter tuning configurations compared to exhaustive search, and should allow near-optimal hyperparameter configurations to be found more efficiently.

    \section{Evaluation}\label{sec:evaluation}

In this section, we evaluate the advancements presented in \cref{sec:impl} to determine the efficacy and efficiency of our hyperparameter tuning method for auto-tuning. 
\Cref{subsec:evaluation_setup} details the experimental setup.
In \cref{subsec:evaluation_results_tuning}, we discuss the efficacy of our method. %
In \cref{subsec:evaluation_results_metastrategies}, we evaluate the performance of meta-strategies to optimize and expand the hyperparameter tuning.
We then evaluate extensive non-exhaustive tuning in \cref{subsec:evaluation_results_extended_tuning}. 
Finally, we evaluate the feasibility and efficiency of our approach in \cref{subsec:evaluation_results_scalability}. %

\subsection{Experimental Setup} \label{subsec:evaluation_setup}

To obtain a diverse set of real-world cases for evaluation, we use the four applications auto-tuned on six GPUs as described in \cref{subsec:impl_fair_data}, resulting in 24 unique search spaces. 
Each application has been fully brute-force auto-tuned on each GPU system as described in \cref{subsec:impl_fair_data}, while all evaluations are performed on the \href{https://www.cs.vu.nl/das/clusters.shtml}{DAS-6 supercomputer} \cite{DASMediumScaleDistributedSystem}. %
The nodes in DAS-6 have a 24-core AMD EPYC-2 7402P CPU, 128 GB of RAM, and are running Rocky Linux 4.18. 
The Python version used is 3.11.7, and \href{https://github.com/KernelTuner/kernel_tuner/releases}{Kernel Tuner at 1.3.0}. 
For a fair evaluation, the hyperparameter tuning is conducted on a training set of twelve search spaces resulting from the four applications on the AMD MI250X, Nvidia A100, and Nvidia A4000, and the test set consists of twelve search spaces resulting from the four applications on the AMD W6600, AMD W7800, and Nvidia A6000. 
These combinations of applications and devices result in a highly diverse set of search spaces as seen in \cite{lurati2024bringing}.

\begin{table}[tb]
    \centering
    \scriptsize
    \caption{Hyperparameter values for the optimization algorithms. Optimal values are in bold, those closest to mean in italics.}
    \label{tab:hyperparams}
    \begin{tabularx}{\linewidth}{|l|r|X|}
        \hline
        \textbf{Algorithm} & \textbf{Hyperparameter} & \textbf{Values} \\
        \hline
        Dual Annealing & method & \{\textbf{COBYLA}, L-BFGS-B, SLSQP, CG, Powell, Nelder-Mead, BFGS, \textit{trust-constr}\} \\
        \hline
        \multirow{4}{*}{Genetic Algorithm} & method & \{\textbf{single\_point}, two\_point, uniform, \textit{disruptive\_uniform}\} \\
        & popsize & \{10, \textbf{\textit{20}}, 30\} \\
        & maxiter & \{\textit{50}, 100, \textbf{150}\} \\
        & mutation\_chance & \{\textbf{5}, 10, \textit{20}\} \\
        \hline
        \multirow{3}{*}{\shortstack{Particle Swarm\\ Optimization (\textit{PSO})}} & popsize & \{10, \textit{20}, \textbf{30}\} \\
        & maxiter & \{\textit{50}, \textbf{100}, 150\} \\
        & c1 & \{\textit{1.0}, 2.0, \textbf{3.0}\} \\
        & c2 & \{\textbf{0.5}, \textit{1.0}, 1.5\} \\
        \hline
        \multirow{4}{*}{Simulated Annealing} & T & \{\textbf{0.5}, \textit{1.0}, 1.5\} \\
        & T\_min & \{0.0001, \textbf{\textit{0.001}}, 0.01\} \\
        & alpha & \{0.9925, \textit{0.995}, \textbf{0.9975}\} \\
        & maxiter & \{\textit{1}, \textbf{2}, 3\} \\
        \hline
    \end{tabularx}
\end{table}

While Kernel Tuner provides a plethora of optimization algorithms as seen in \cref{tab:autotuners}, due to space constraints, we select the four optimization algorithms shown in \cref{tab:hyperparams} to represent the diverse range of global optimization algorithms and hyperparameters available. %
\textit{Simulated Annealing} explores the search space by probabilistically accepting worse solutions to escape local optima, gradually reducing this randomness over time. 
\textit{Dual Annealing} combines elements of simulated annealing with local search.  
\textit{Genetic Algorithm} evolves solutions over generations using selection, crossover, and mutation operators.
Particle swarm optimization (\textit{PSO}) navigates optimization landscapes through information sharing among candidate solutions.
A sensitivity test of the hyperparameters using the non-parametric Kruskal-Wallis test and mutual information scoring revealed that the $W$ hyperparameter of PSO had no meaningful effect on the score, and as such is left out.
While several of the hyperparameters used in \cref{tab:hyperparams} are numerical and thus have a much larger set of possible values, we limit the values to those shown in order to conduct an exhaustive search of all possible combinations. 
This allows for determining the potential performance impact of hyperparameter tuning and enables a fair evaluation of meta-strategies on this exhaustive data. 
Each combination of hyperparameter values shown in \cref{tab:hyperparams} is run 25 times on each of the 12 search spaces to obtain a robust performance score. This means that tuning the hyperparameters of e.g. Genetic Algorithm as in \cref{tab:hyperparams} requires running the algorithm 32400 times. %
The allocated budget for each run is equivalent to the time it takes the baseline to reach 95\% of the distance between the search space median and optimum, as discussed in \cref{subsec:impl_hyperparameter_tuning}. 
In the following performance comparisons, each instance is re-executed with 100 repeats to mitigate stochasticity.

\subsection{Hyperparameter Tuning Efficacy} \label{subsec:evaluation_results_tuning}
\begin{figure}[tbp]
    \centering
    \includegraphics[width=0.99\linewidth]{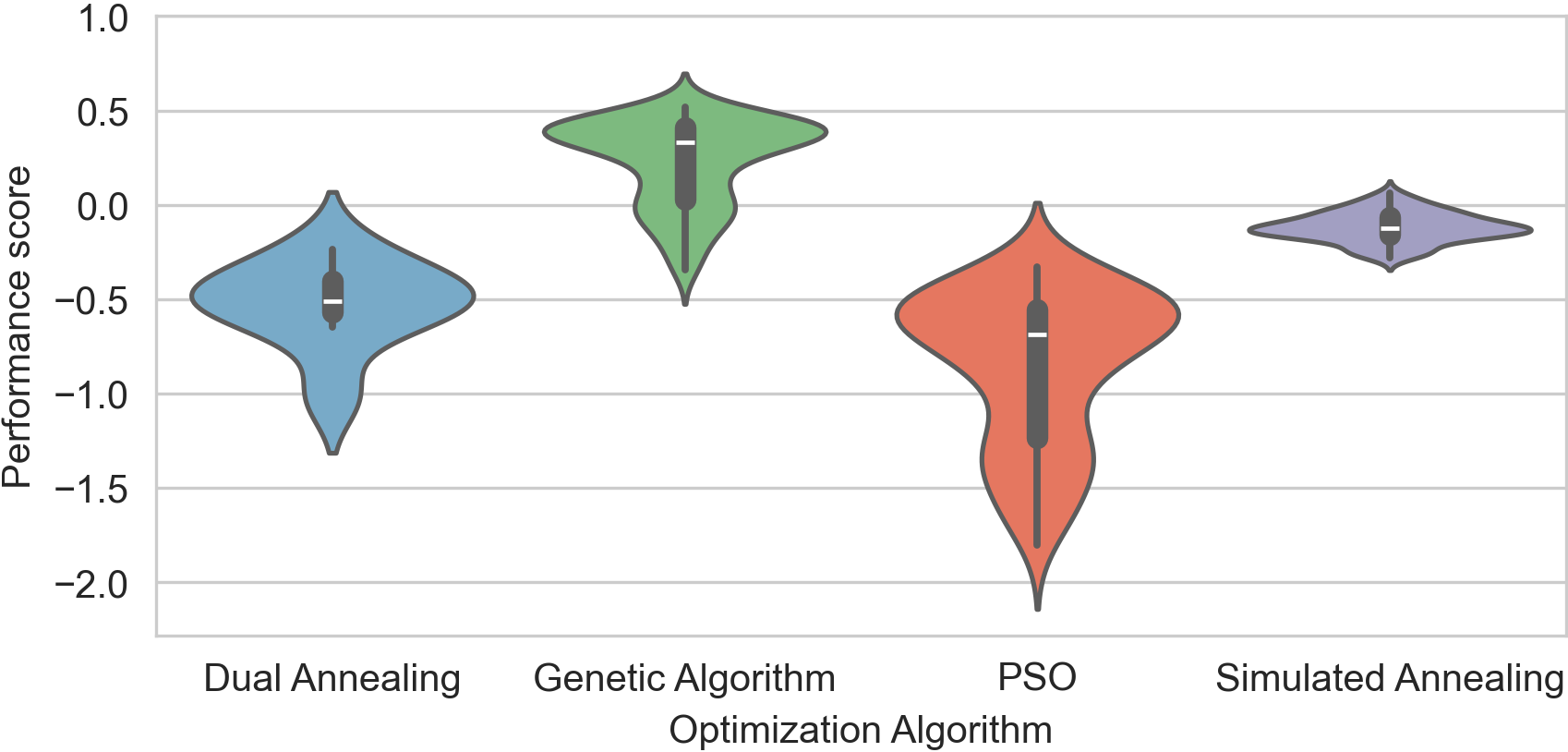}
    \vspace{-0.5cm}
    \caption{Violin plots of the performance scores (higher is better) for all hyperparameter configurations of the evaluated optimization algorithms, showing the mean (white line), boxplot (black box), and distribution (area).}
    \label{fig:results_tuning_violin_plot}
\end{figure}

First, we discuss the results regarding the impact and generalization of our hyperparameter tuning method. 

\Cref{fig:results_tuning_violin_plot} depicts the distribution of hyperparameter configuration performance scores for each optimization algorithm. 
With these scores, which will be used throughout this evaluation, an algorithm with a score of 0 performs as well as the baseline, a negative indicates worse performance than the baseline, and a score of 1 indicates that the optimum is consistently found immediately - the ultimate goal of any optimization algorithm. 
For more information on the performance scores, see \cref{subsec:impl_hyperparameter_tuning}. 
The violin plots in \cref{fig:results_tuning_violin_plot} depict large differences in scores between the various hyperparameter configurations of each optimization algorithm. In addition, it shows the difference in sensitivity to hyperparameter tuning between optimization algorithms; e.g. PSO is substantially more sensitive than simulated annealing. 
An average difference in performance score of 0.865 between the best and worst hyperparameter configurations highlights the substantial performance improvement potentially gained. %

\begin{figure}[tbp]
    \centering
    \includegraphics[width=0.99\linewidth]{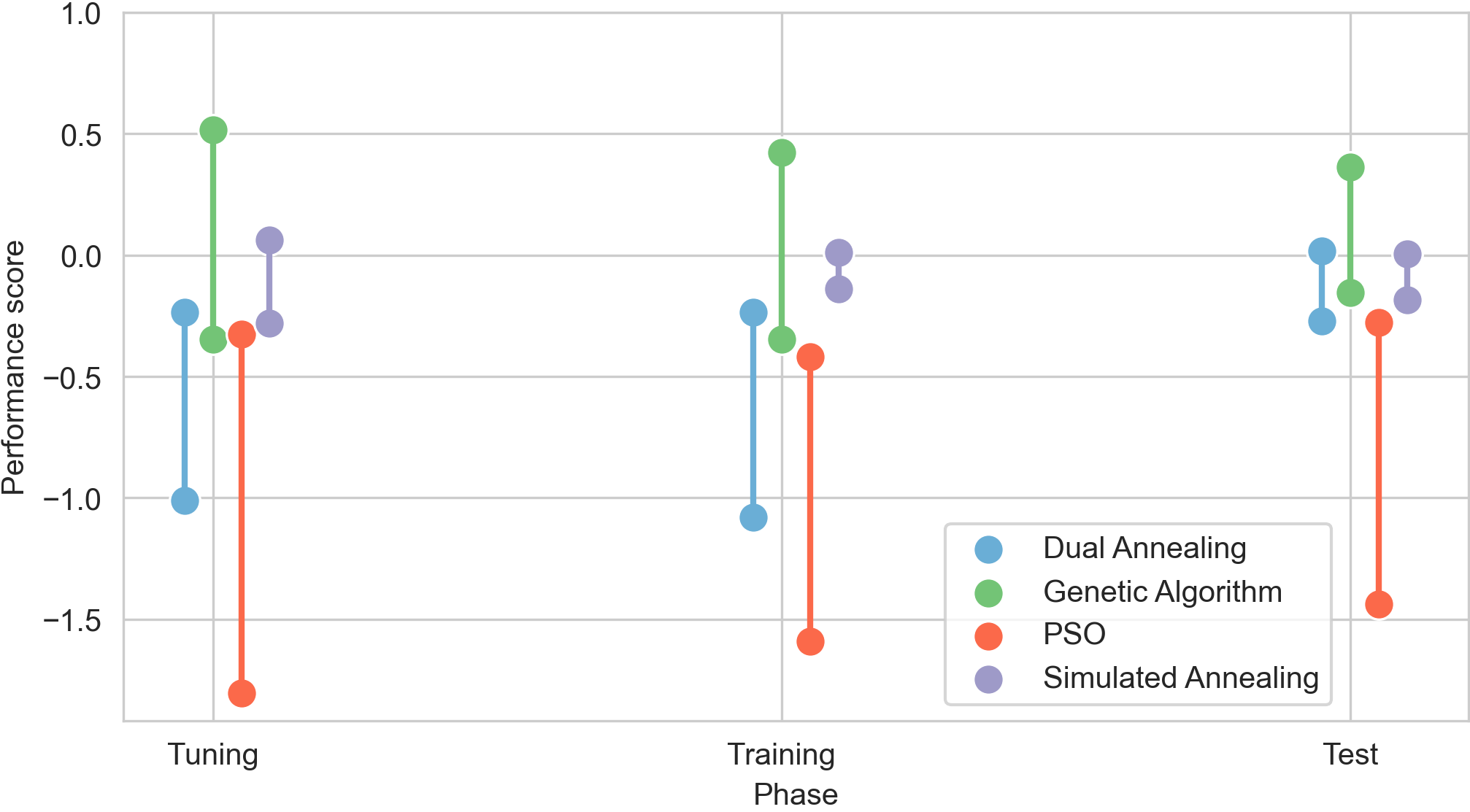}
    \caption{Best and worst scores on tuning, training (re-executed for comparison), and test for evaluated optimization algorithms.}
    \label{fig:results_tuning_training_test}
\end{figure}

\begin{figure}[tbp]
\centering
\subfloat[Dual Annealing suboptimal\label{fig:results_hyperparameter_heatmap_da_inv_tuned}]{%
  \includegraphics[width=0.499\linewidth]{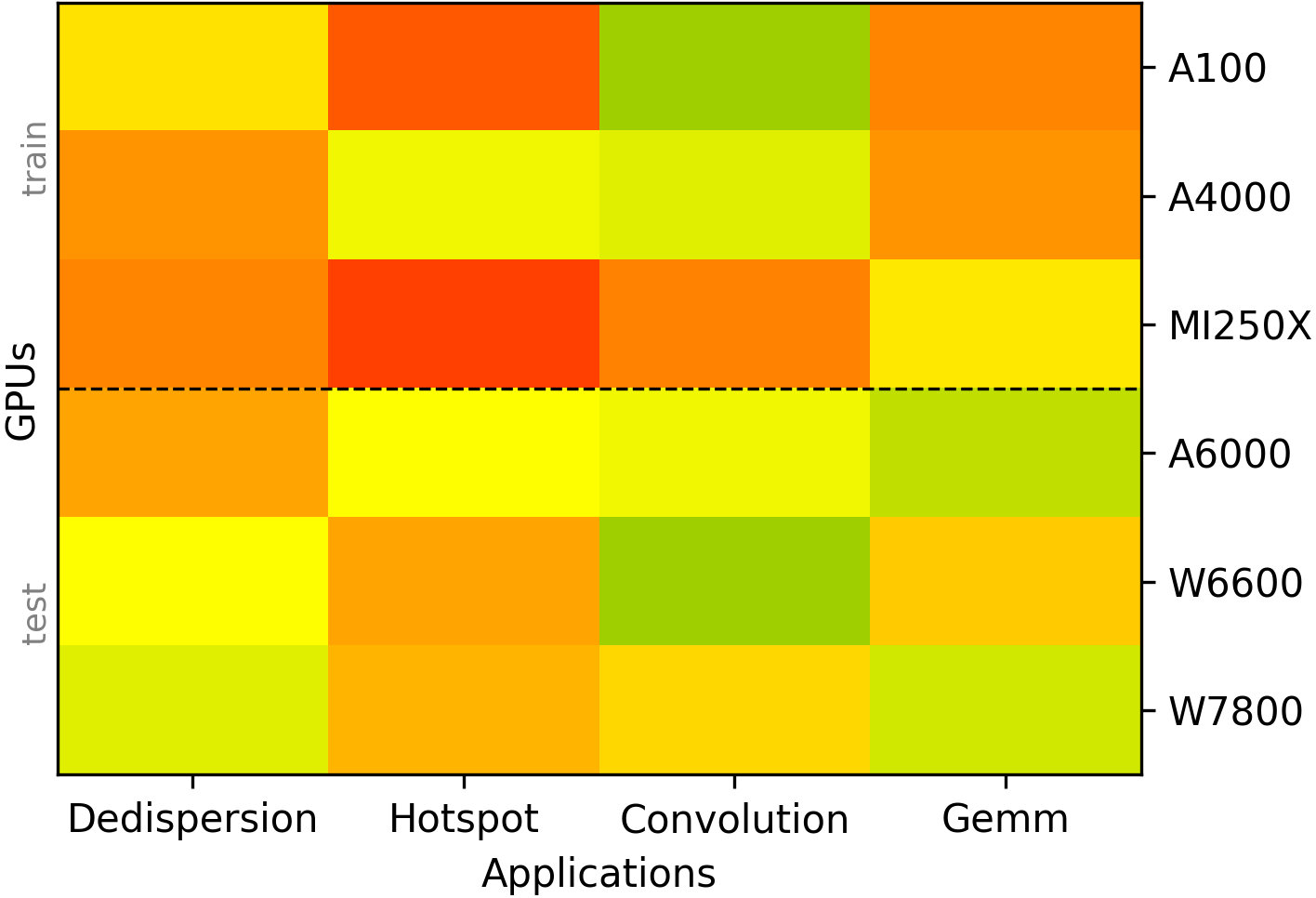}%
}\hfil
\subfloat[Dual Annealing optimal\label{fig:results_hyperparameter_heatmap_da_tuned}]{%
  \includegraphics[width=0.499\linewidth]{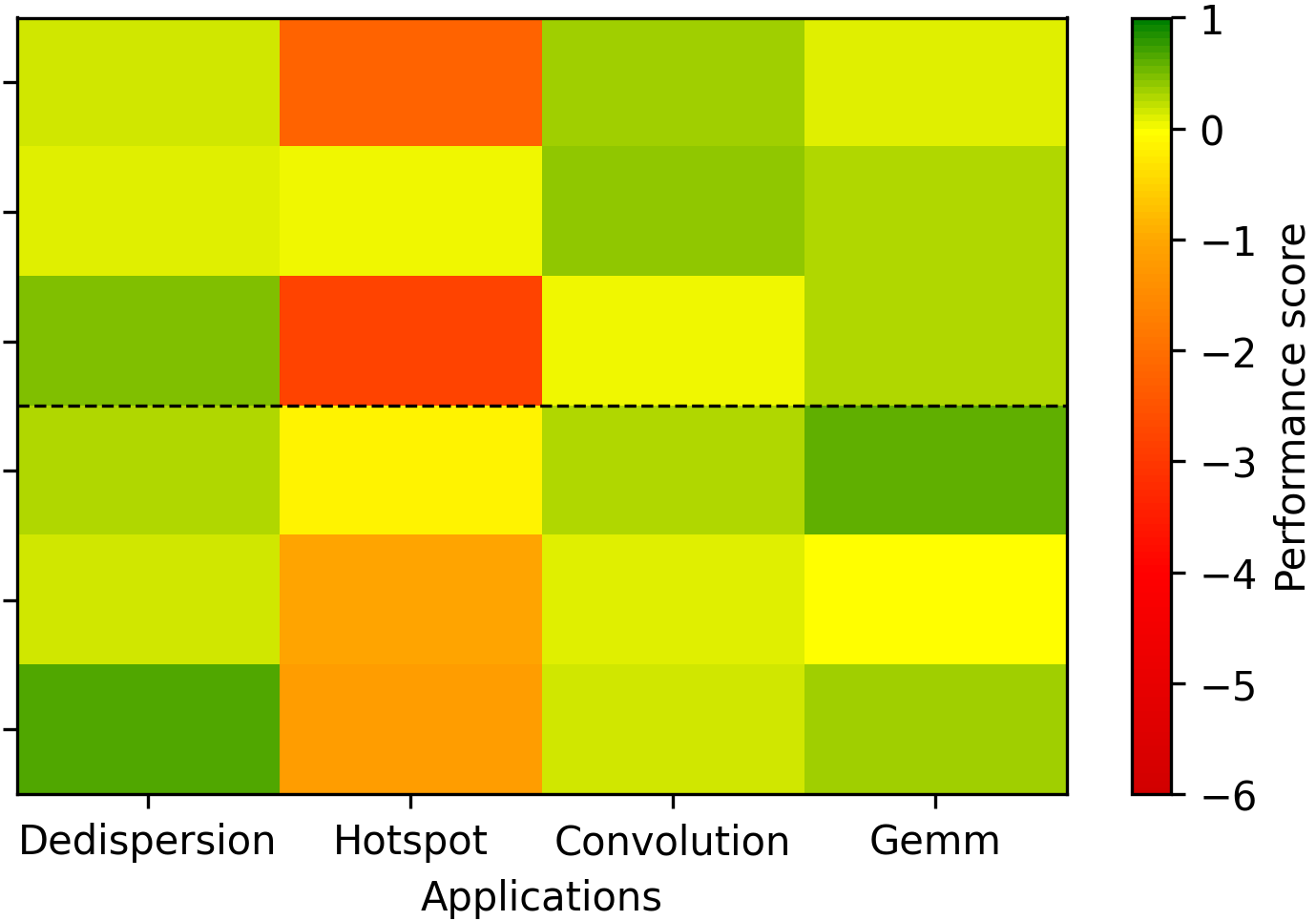}%
}
\\
\subfloat[Genetic Algorithm suboptimal\label{fig:results_hyperparameter_heatmap_ga_inv_tuned}]{%
  \includegraphics[width=0.499\linewidth]{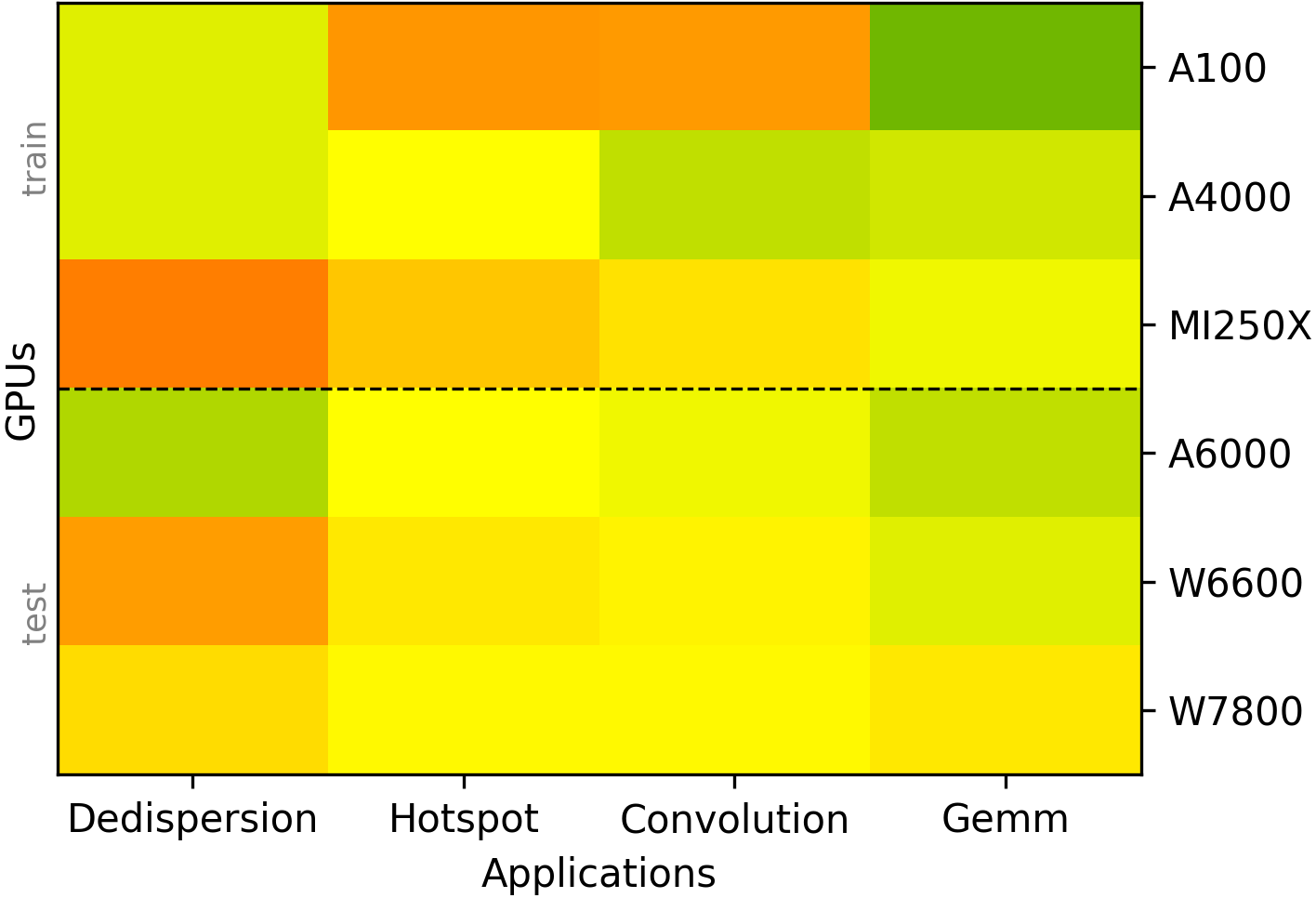}%
}\hfil
\subfloat[Genetic Algorithm optimal\label{fig:results_hyperparameter_heatmap_ga_tuned}]{%
  \includegraphics[width=0.499\linewidth]{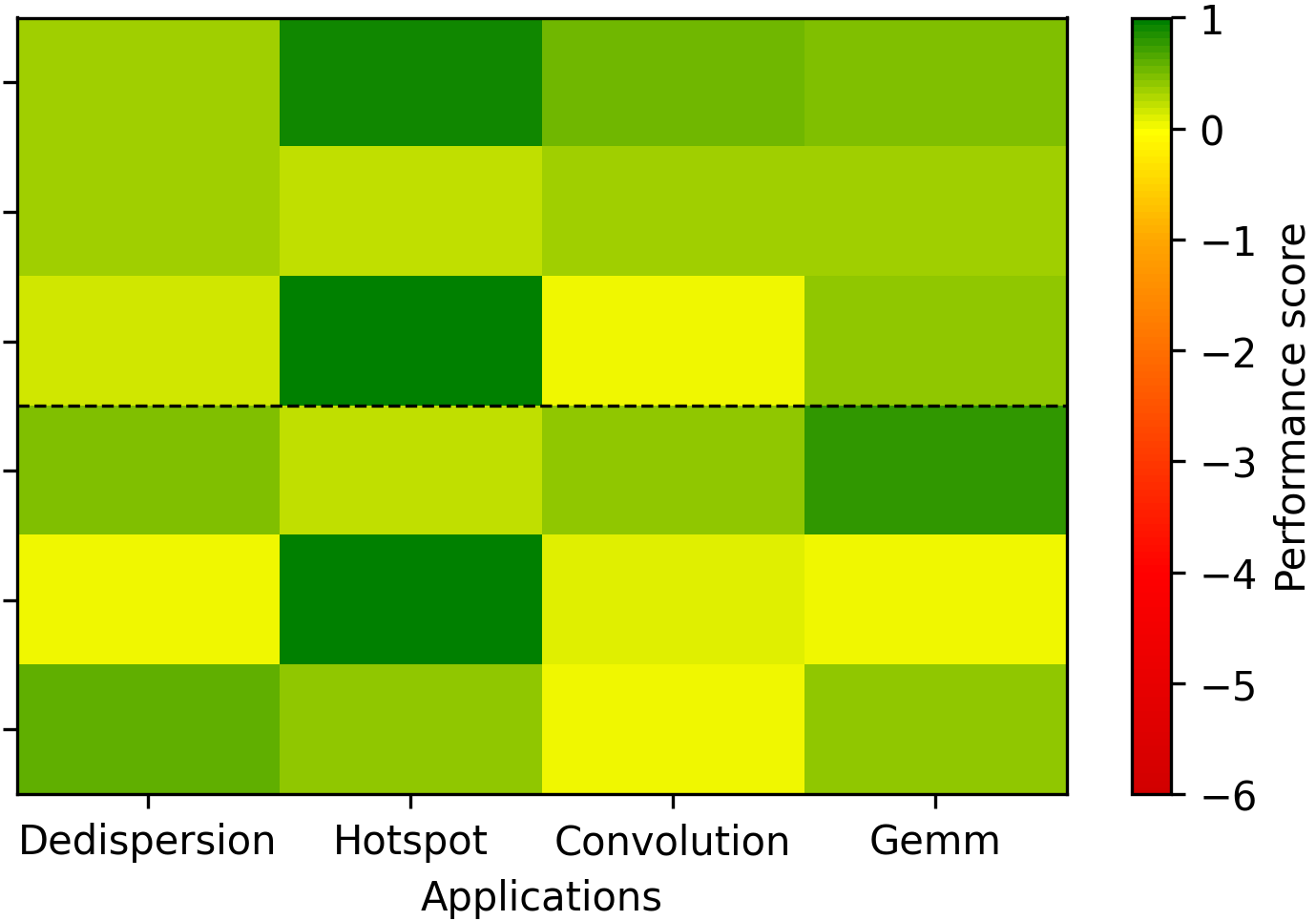}%
}
\\
\subfloat[PSO suboptimal\label{fig:results_hyperparameter_heatmap_pso_inv_tuned}]{%
  \includegraphics[width=0.499\linewidth]{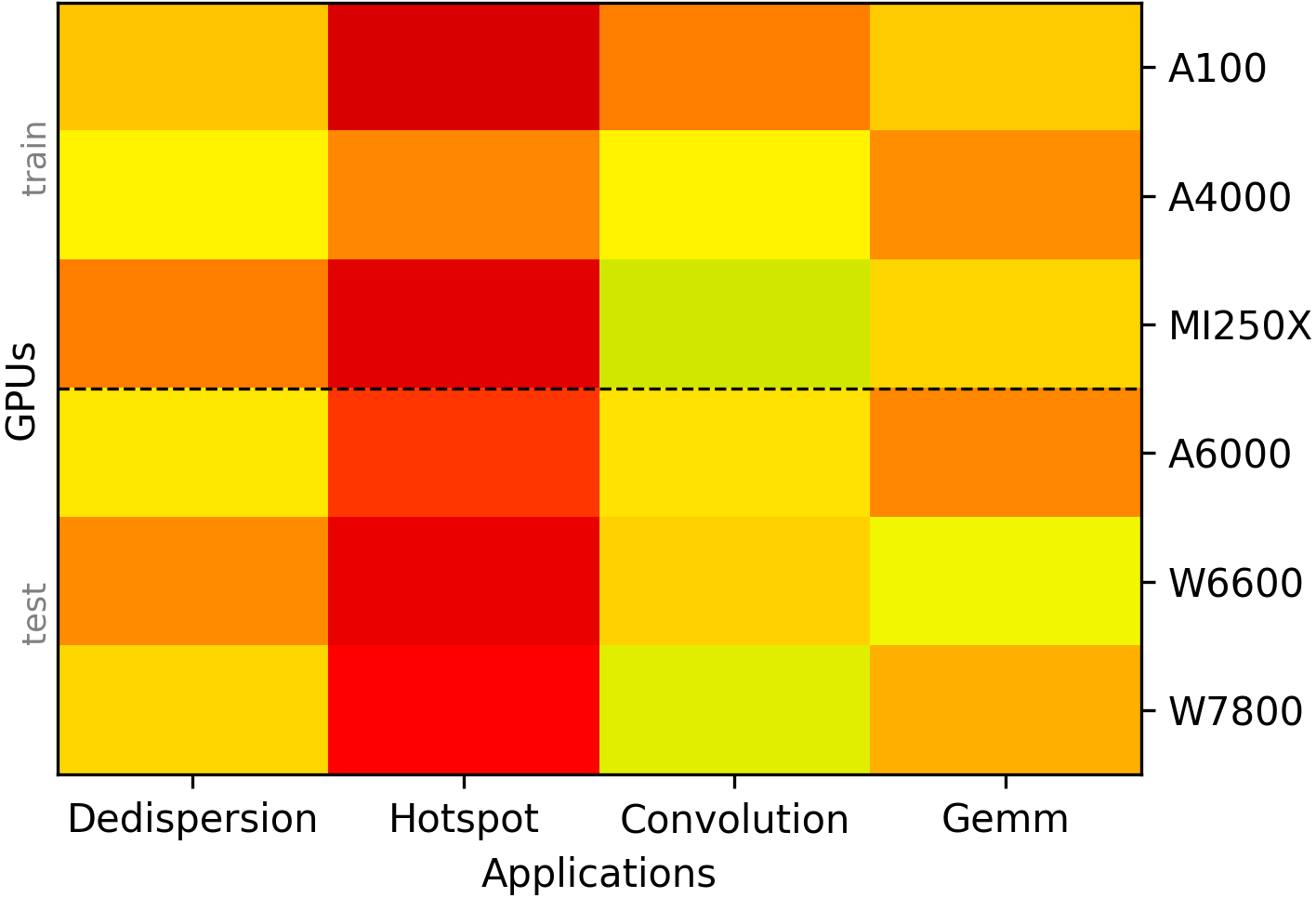}%
}\hfil
\subfloat[PSO optimal\label{fig:results_hyperparameter_heatmap_pso_tuned}]{%
  \includegraphics[width=0.499\linewidth]{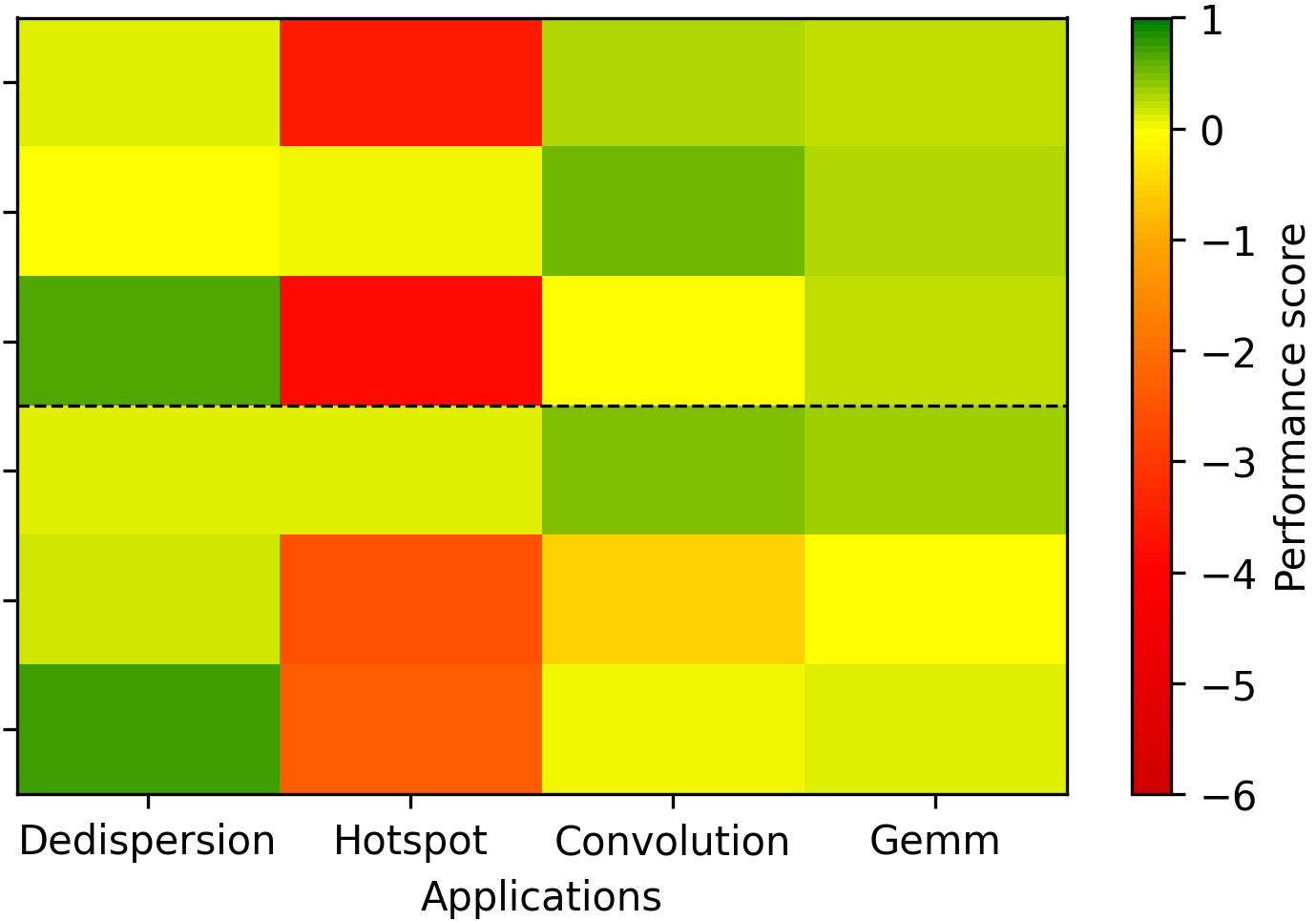}%
}
\\
\subfloat[Simulated Annealing suboptimal\label{fig:results_hyperparameter_heatmap_sa_inv_tuned}]{%
  \includegraphics[width=0.499\linewidth]{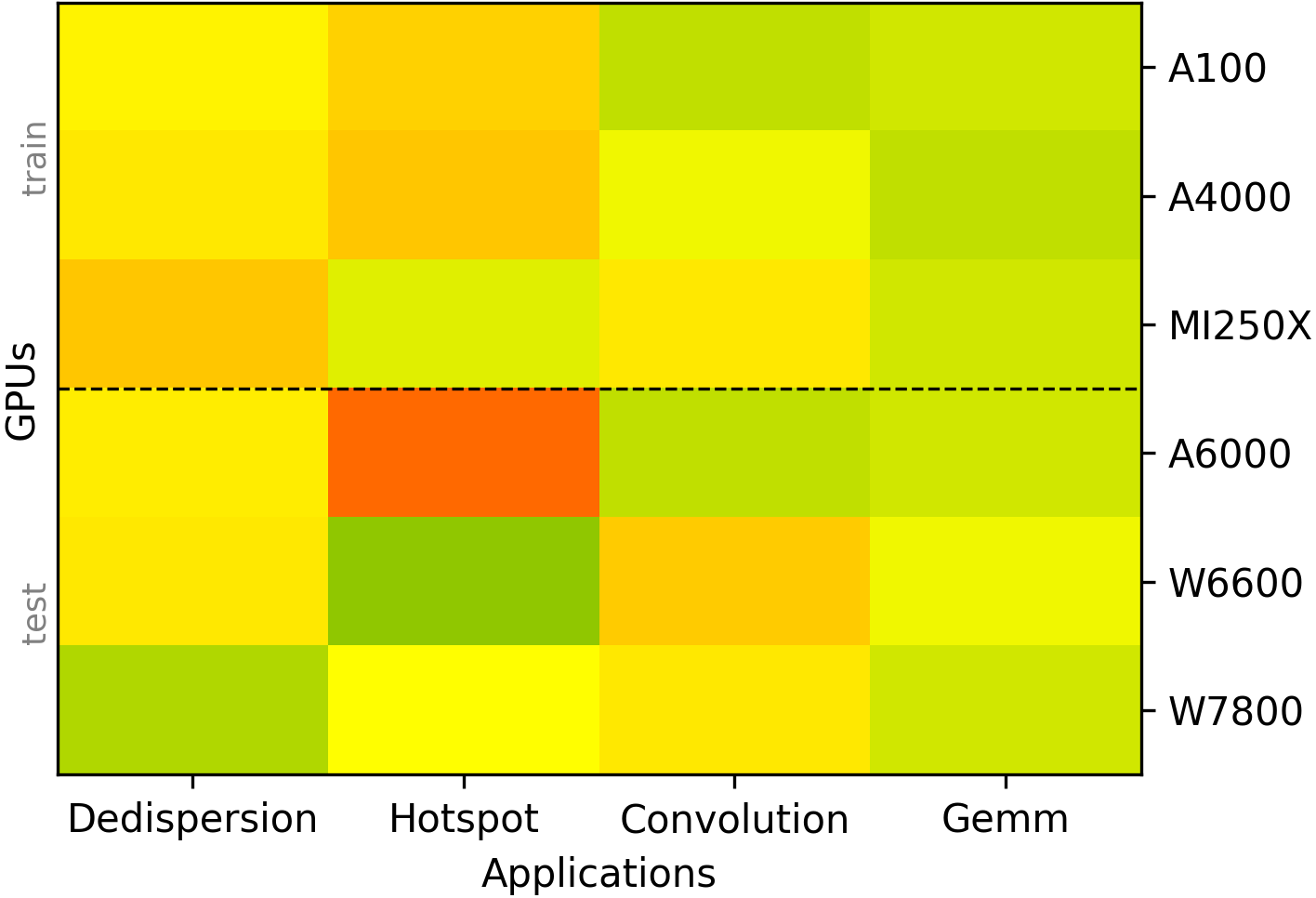}%
}\hfil
\subfloat[Simulated Annealing  optimal\label{fig:results_hyperparameter_heatmap_sa_tuned}]{%
  \includegraphics[width=0.499\linewidth]{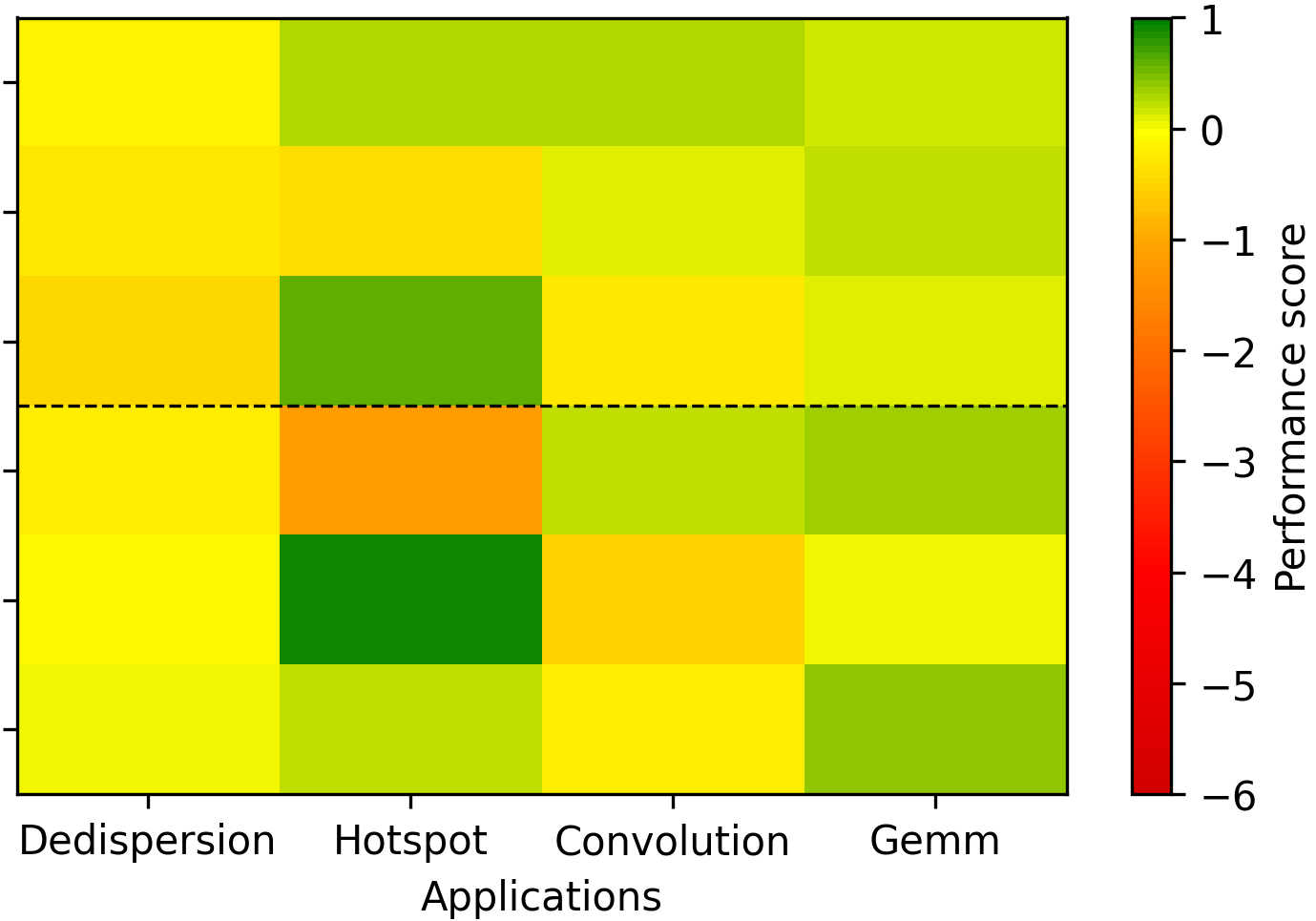}%
}
\caption{Impact of tuning on optimization algorithm performance; left-hand and right-hand columns show the performance on all search spaces for suboptimal and optimal optimization algorithm versions, respectively.}
\label{fig:results_hyperparameter_heatmap_per_searchspace}
\end{figure}

We validate this by examining the overall performance and generalization of the hyperparameter tuning across the search spaces. 
\Cref{fig:results_tuning_training_test} shows that the performance scores of the best- and worst-performing hyperparameter configurations obtained in tuning, which were executed 25 times during tuning, remain largely stable when re-executed with 100 repeats on the same training data. 
Most importantly, the best configuration performance is also comparable on the test set of search spaces not tuned on, indicating excellent generalization performance. 

To further verify the generalization performance, we can compare the performance on the 24 individual search spaces. 
In \cref{fig:results_hyperparameter_heatmap_per_searchspace}, visual inspection of both the suboptimal (worst performing, left column) and optimal (best performing, right column) versions of each optimization algorithm reveals that each of the optimal versions improves upon their counterpart in general, instead of over-optimizing on a limited number of search spaces to boost the score. 
Interestingly, all optimization algorithms, except for the optimal versions of \textit{genetic algorithm} and \textit{simulated annealing}, struggle with the hotspot application search spaces. 
Most importantly, it can be seen that for the vast majority of search spaces in both the training (upper half of each plot) and test sets (lower half of each plot), the performance is substantially improved. 

\begin{figure}[tb]
    \centering
    \includegraphics[width=0.99\linewidth]{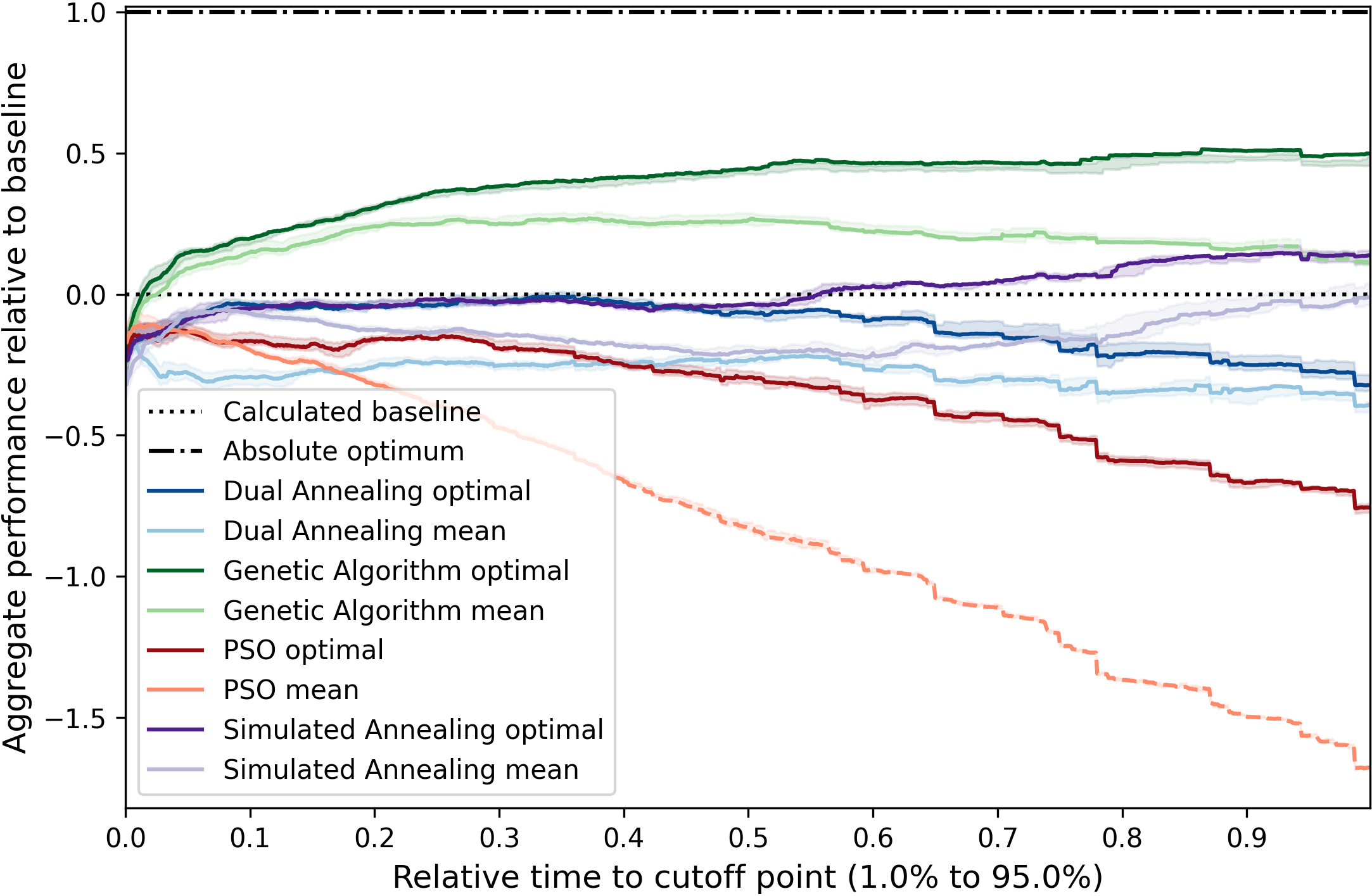}
    \vspace{-0.8cm}
    \caption{Aggregate performance over time between optimization algorithms with mean and optimal hyperparameters across all search spaces.}
    \label{fig:results_aggregate}
\end{figure}

Finally, we compare the optimal hyperparameter configuration to the most average-performing hyperparameter configuration (i.e., closest to the mean score in \cref{fig:results_tuning_violin_plot}) to gauge the average impact of hyperparameter tuning. 
\Cref{fig:results_aggregate} compares these optimal and average configurations of each optimization algorithm over time across all the search spaces, showing that the optimal optimization algorithms outperform their average counterparts by a wide margin. 
Quantifying this difference with the performance score, Dual Annealing is improved by 0.170, Genetic Algorithm by 0.192, PSO by 0.473, and Simulated Annealing by 0.149, for an average improvement of 94.8\% over the average hyperparameter configuration. %

\begin{figure}[tbph]
    \centering
    \includegraphics[width=0.97\linewidth]{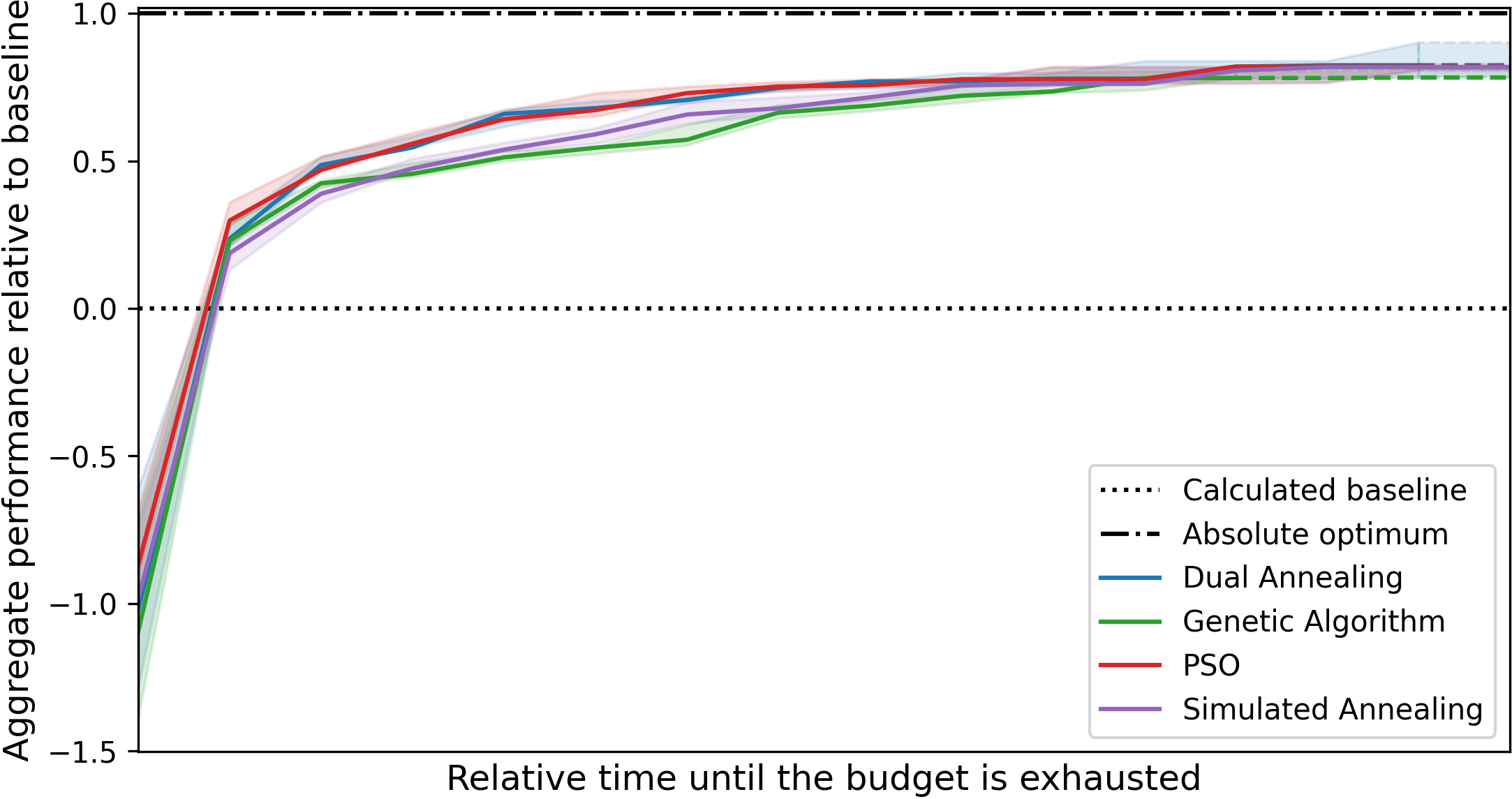}
    \vspace{-0.3cm}
    \caption{Aggregate performance over time for various meta-strategies on the hyperparameter tuning search spaces, 100 repeated runs each.}
    \label{fig:results_aggregate_metastrategy}
\end{figure}

\subsection{Meta-strategies for Hyperparameter Tuning} \label{subsec:evaluation_results_metastrategies}
With the efficacy of hyperparameter tuning established, we evaluate the usage of meta-strategies to efficiently explore a more expansive hyperparameter search space as opposed to exhaustive hyperparameter tuning. 

By running various meta-strategies on the exhaustively evaluated hyperparameter tuning spaces for each of the algorithms and applying the performance scoring once more, we can investigate whether hyperparameter configurations in auto-tuning can be effectively optimized with optimization algorithms themselves. 
These meta-strategies can be the same optimization strategies that are already included with Kernel Tuner; for simplicity, we will reuse the optimization algorithms tuned earlier as meta-strategies. 
\Cref{fig:results_aggregate_metastrategy} depicts the results, with all optimization algorithms performing very well after an initial startup cost, with an average performance score of 0.223. 
These findings demonstrate the substantial increase in efficiency of using a meta-strategy over random or exhaustive search, which allows much wider ranges of hyperparameter values to be tuned with a high probability that a near-optimal hyperparameter configuration is found. 

\subsection{Extended Hyperparameter Tuning} \label{subsec:evaluation_results_extended_tuning}
\begin{table}[tb]
    \centering
    \scriptsize
    \caption{Extended hyperparameter values for the optimization algorithms. Optimal values are in bold.}
    \label{tab:hyperparams-extended}
    \begin{tabularx}{\linewidth}{|l|r|X|}
        \hline
        \textbf{Algorithm} & \textbf{Hyperparameter} & \textbf{Values} \\
        \hline
        \multirow{4}{*}{Genetic Algorithm} & method & \{\textbf{single\_point}, two\_point, uniform, disruptive\_uniform\} \\
        & popsize & \{ $2,\; 4,\; 6,\,\dots,\textbf{26},\;\dots,\,50$ \} \\
        & maxiter & \{ $10,\; 20,\; 30,\,\dots,\textbf{150},\;\dots,\,200$\} \\
        & mutation\_chance & \{ $\textbf{5},\; 10,\; 15,\,\dots,\,100$ \} \\
        \hline
        \multirow{3}{*}{\shortstack{Particle Swarm\\ Optimization (\textit{PSO})}} & popsize & \{ $2,\; 4,\; 6,\,\dots,\,\textbf{50}$ \} \\
        & maxiter & \{ $10,\; 20,\; 30,\,\dots,\,\textbf{190},\;200$ \} \\
        & c1 & \{ $1.0,\; 1.25,\; 1.5,\,\dots,\,\textbf{3.5}$ \} \\
        & c2 & \{ $0.5,\; 0.75,\; \textbf{1.0},\,\dots,\,2.0$ \} \\
        \hline
        \multirow{4}{*}{Simulated Annealing} & T & \{ $\textbf{0.1},\; 0.2,\; 0.3,\,\dots,\,2.0$ \} \\
        & T\_min & \{ $\textbf{0.0001},\; 0.0011,\,\dots,\,0.1$ \} \\
        & alpha & \{0.9925, 0.995, \textbf{0.9975}\} \\
        & maxiter & \{ $\textbf{1},\; 2,\; 3,\,\dots,\,10$ \} \\
        \hline
    \end{tabularx}
\end{table}

\begin{figure}[tbp]
\centering
\subfloat[Genetic Algorithm average\label{fig:results_hyperparameter_heatmap_ga_inv_tuned_extended}]{%
  \includegraphics[width=0.499\linewidth]{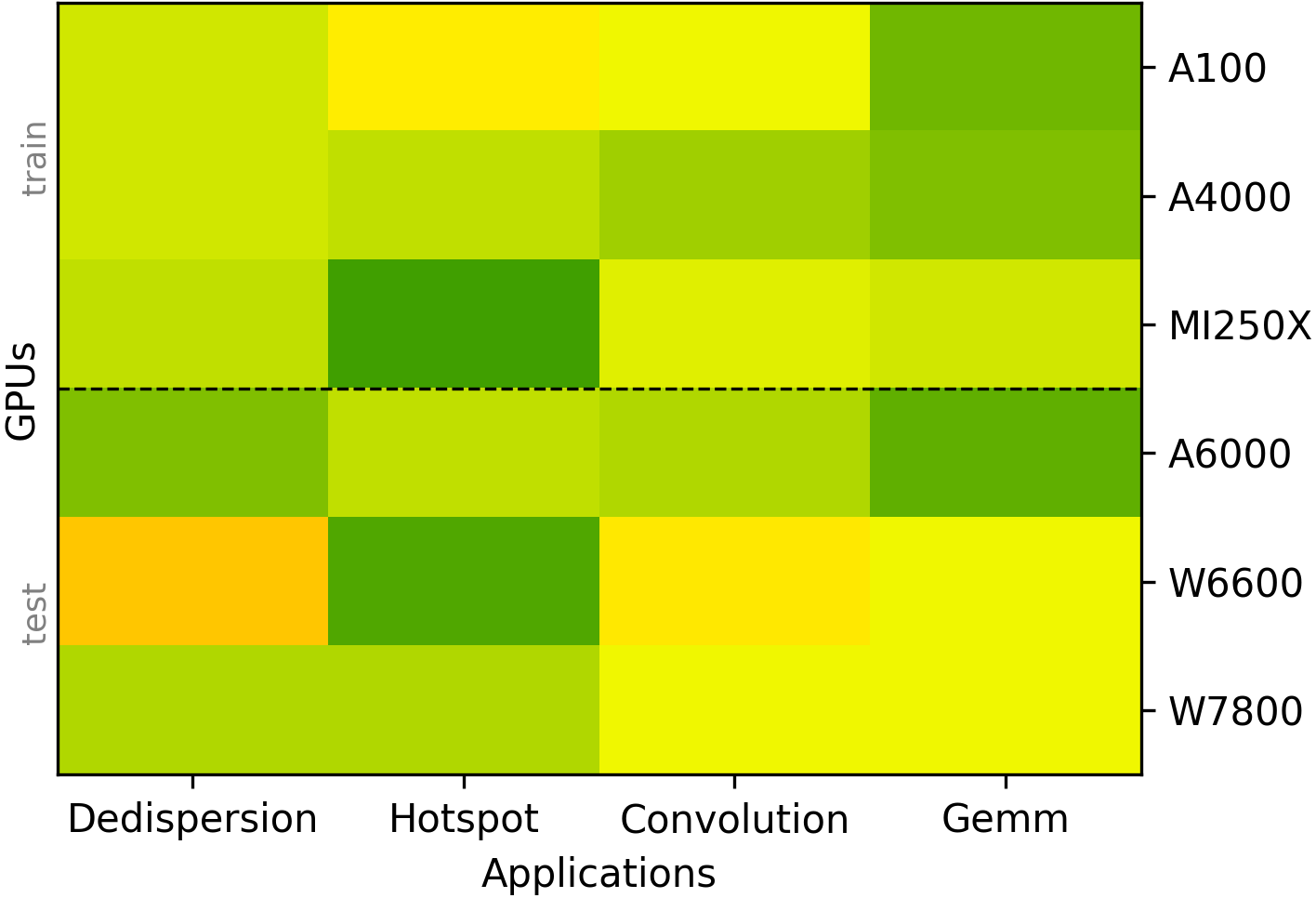}%
}\hfil
\subfloat[Genetic Algorithm optimal\label{fig:results_hyperparameter_heatmap_ga_tuned_extended}]{%
  \includegraphics[width=0.499\linewidth]{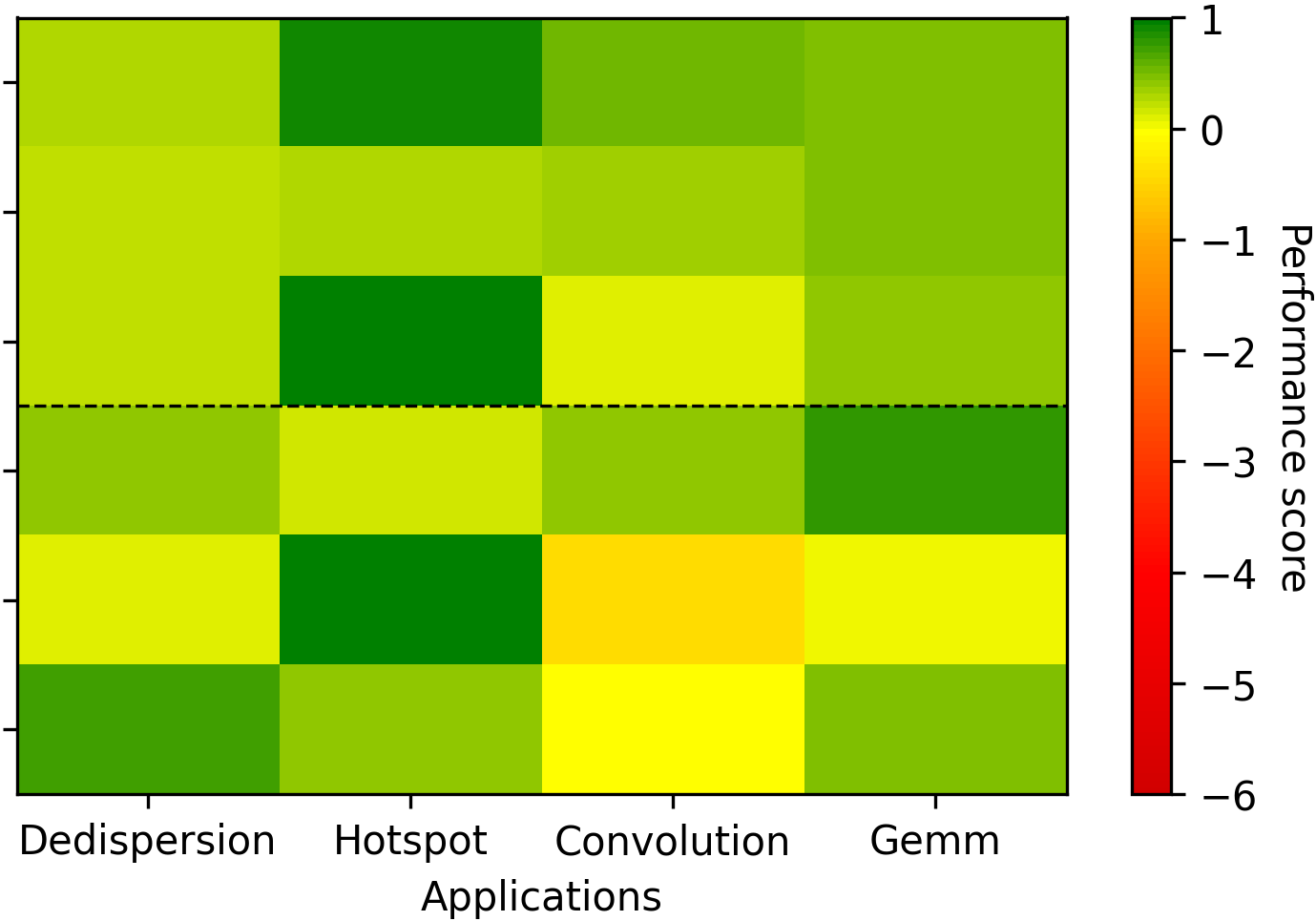}%
}
\\
\subfloat[PSO average\label{fig:results_hyperparameter_heatmap_pso_inv_tuned_extended}]{%
  \includegraphics[width=0.499\linewidth]{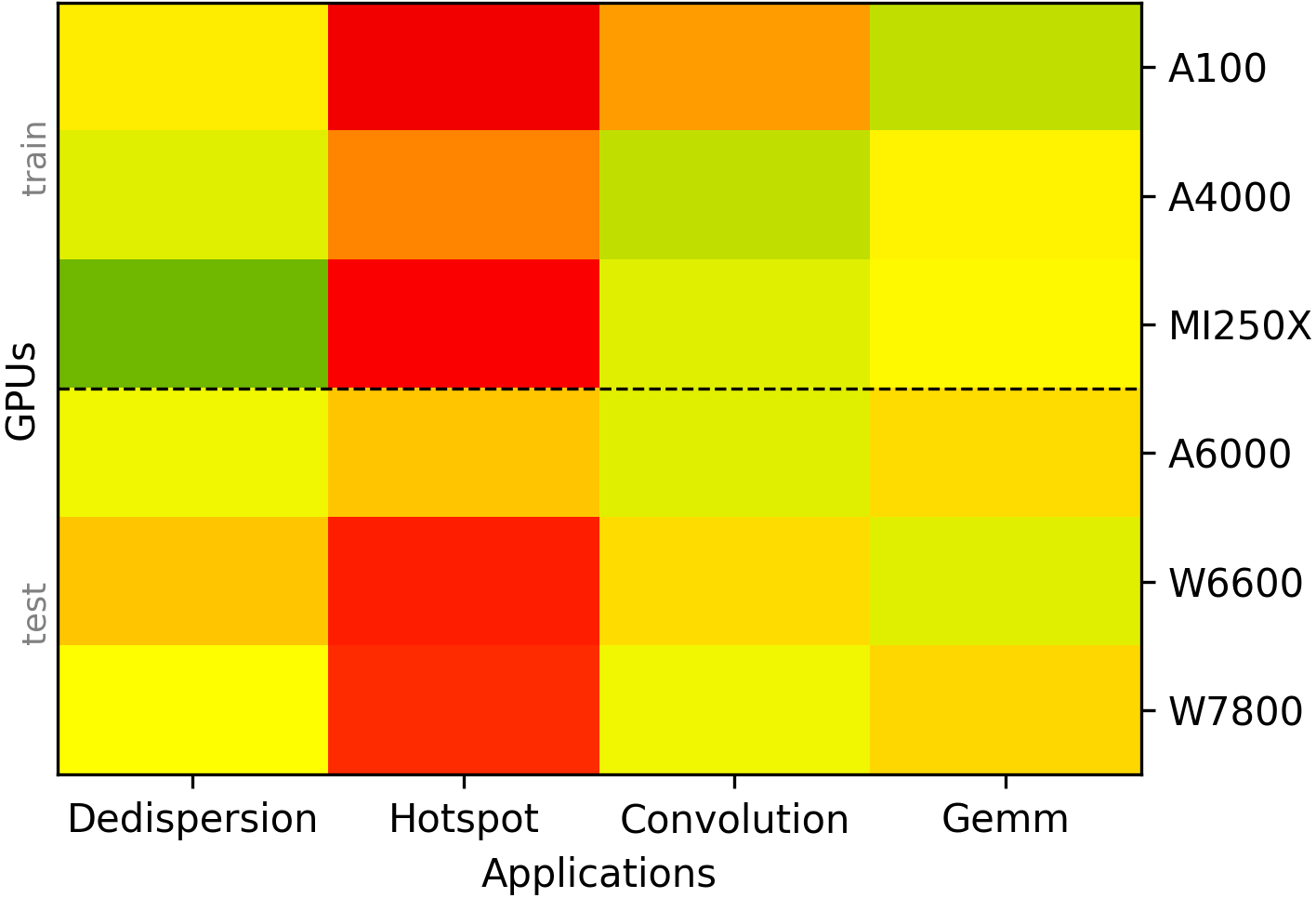}%
}\hfil
\subfloat[PSO optimal\label{fig:results_hyperparameter_heatmap_pso_tuned_extended}]{%
  \includegraphics[width=0.499\linewidth]{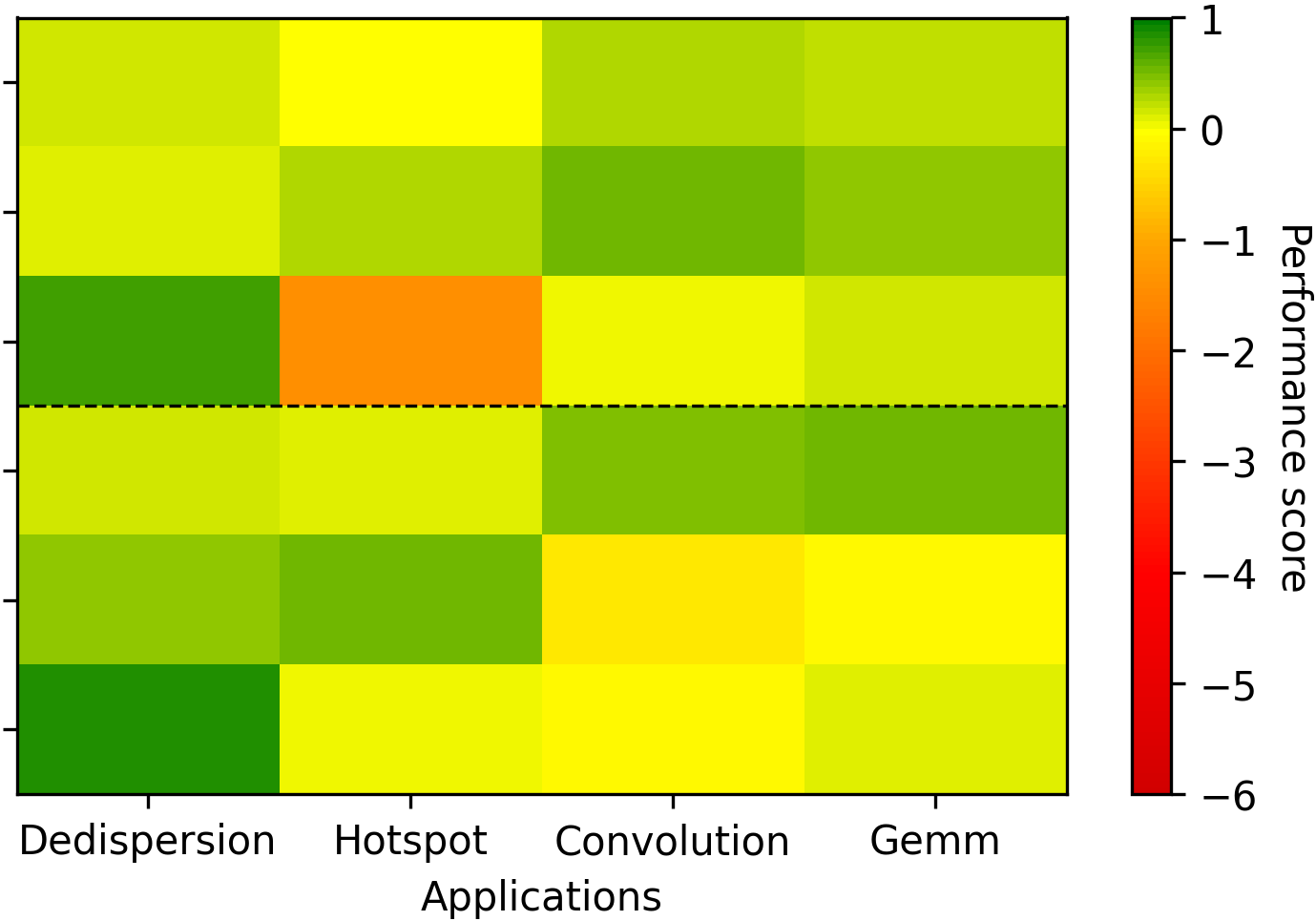}%
}
\\
\subfloat[Simulated Annealing average\label{fig:results_hyperparameter_heatmap_sa_inv_tuned_extended}]{%
  \includegraphics[width=0.499\linewidth]{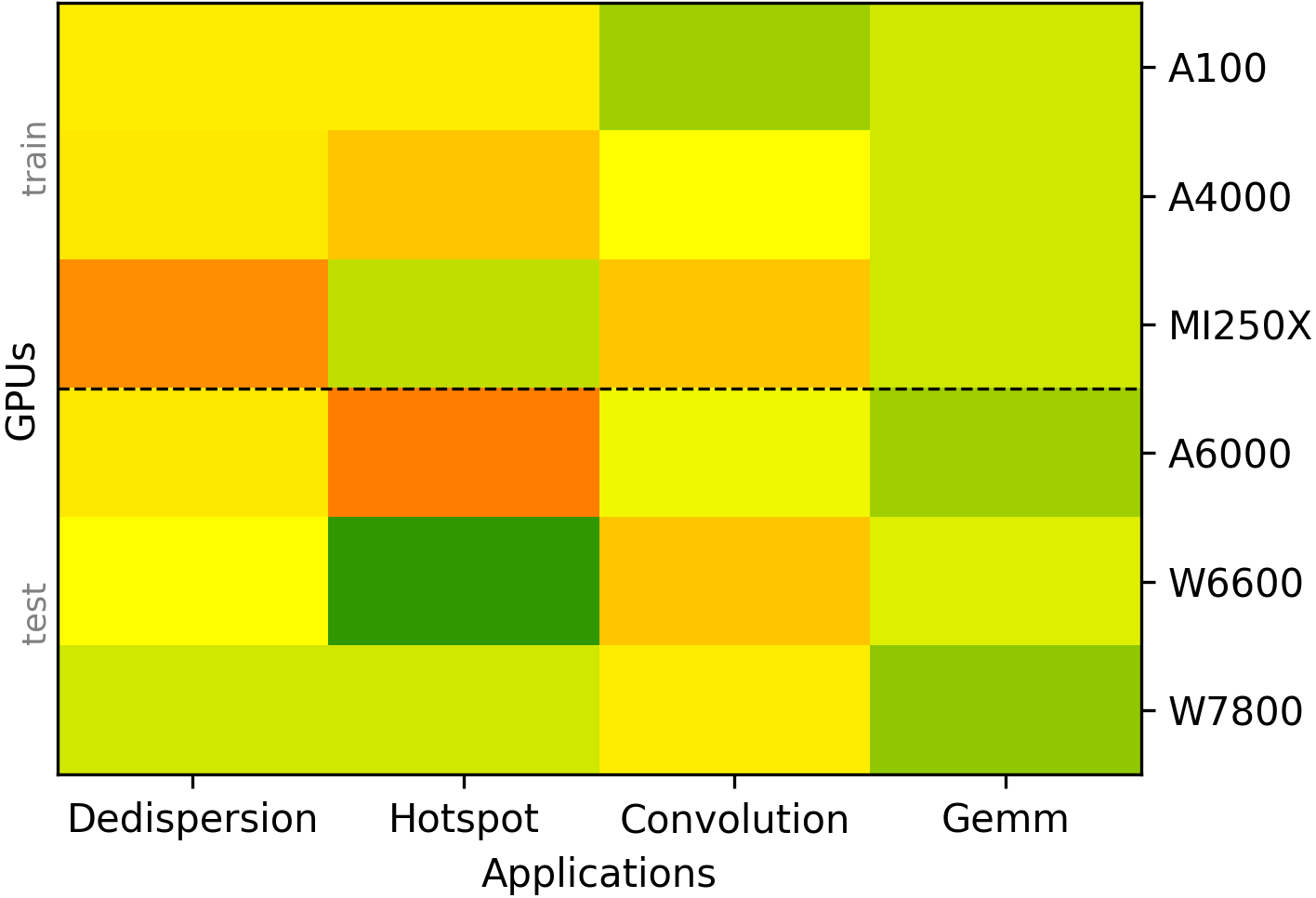}%
}\hfil
\subfloat[Simulated Annealing  optimal\label{fig:results_hyperparameter_heatmap_sa_tuned_extended}]{%
  \includegraphics[width=0.499\linewidth]{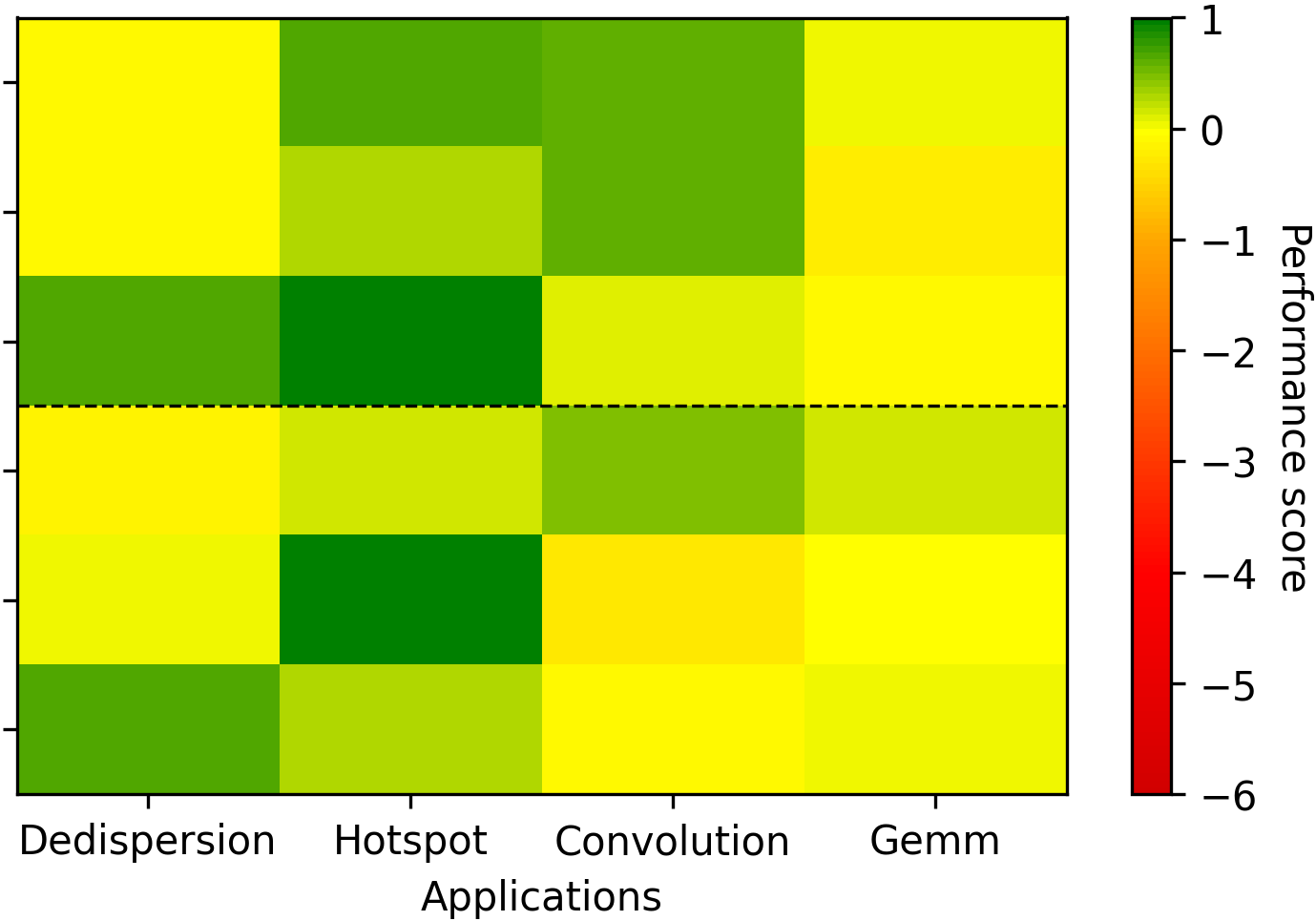}%
}
\caption{Impact of extended tuning on optimization algorithm performance; left-hand and right-hand columns show the performance on all search spaces for average and optimal optimization algorithm versions, respectively.}
\label{fig:results_hyperparameter_heatmap_per_searchspace_extended}
\end{figure}

\begin{figure}[tb]
    \centering
    \includegraphics[width=0.99\linewidth]{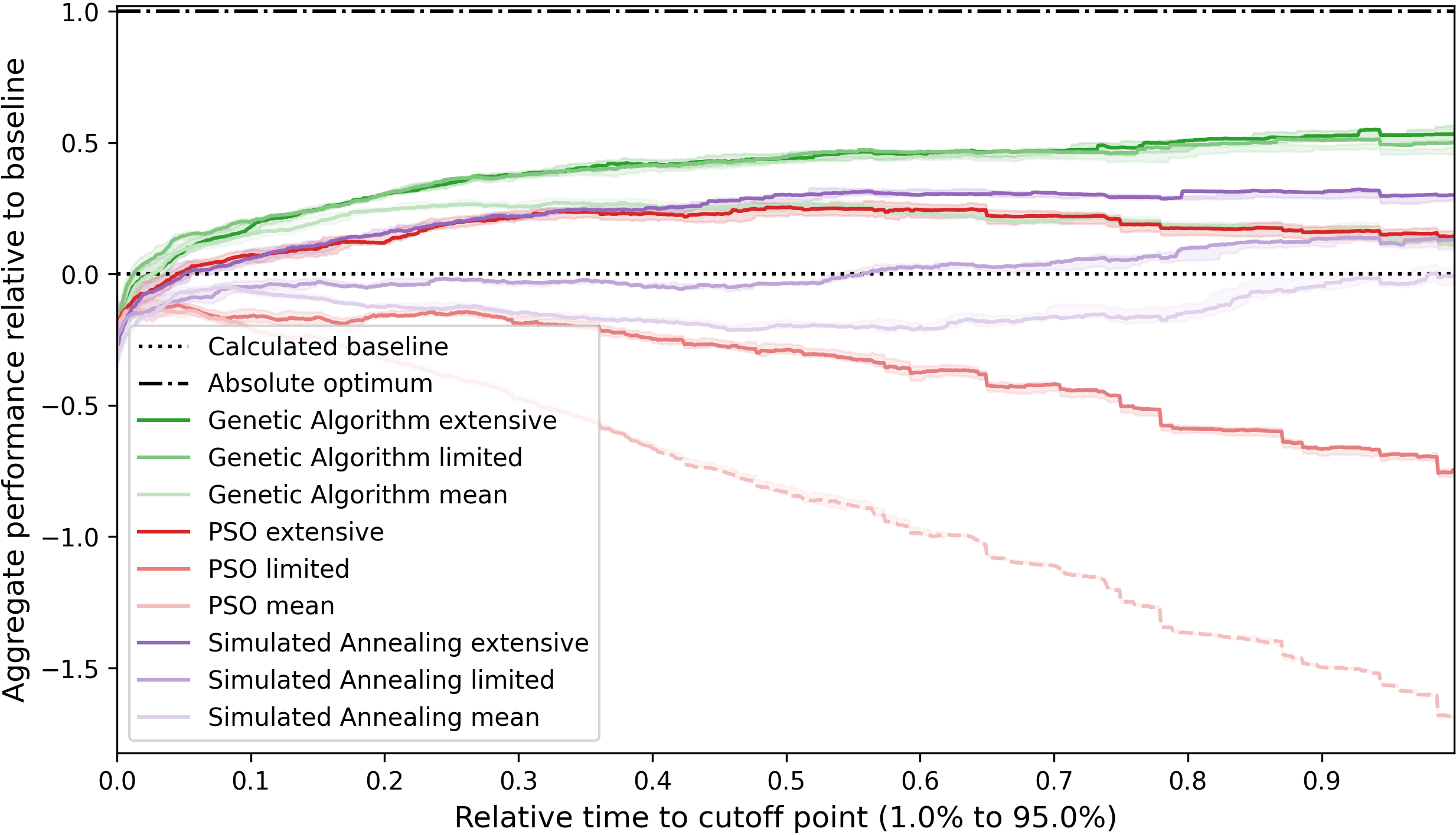}
    \vspace{-0.8cm}
    \caption{Aggregate performance over time between optimization algorithms with mean and optimal (limited and extended) hyperparameters across all search spaces.}
    \label{fig:results_aggregate_extended}
\end{figure}

Having established that hyperparameter tuning auto-tuning optimization algorithms generalizes well and is effective, and meta-strategies are efficient at finding near-optimal hyperparameter configurations, we perform a final extensive, non-exhaustive hyperparameter tuning on the optimization algorithms with numerical hyperparameters exposed for a more realistic approach to hyperparameter tuning. 
This extensive hyperparameter tuning is ran on each optimization algorithm for 7 days using Dual Annealing as a meta-strategy. 
The hyperparameter values are shown in \cref{tab:hyperparams-extended}. 

We can compare the optimal hyperparameter configuration of this extended tuning to the most average-performing hyperparameter configuration of \cref{subsec:evaluation_results_tuning} to gauge the average impact of a realistic hyperparameter tuning scenario. 
The improvements between the average configuration of the limited tuning and the optimal configuration of the extended tuning, seen for each search space in \cref{fig:results_hyperparameter_heatmap_per_searchspace_extended}, indicate that the hyperparameter tuning enhances overall performance (on both train and test), like before in \cref{fig:results_hyperparameter_heatmap_per_searchspace}. 
\Cref{fig:results_aggregate_extended} compares the optimal and average configurations of each optimization algorithm over time across all the search spaces, showing that after the optimization algorithms with extended tuning outperform their exhaustive counterparts by an even wider margin than previously with the limited exhaustive hyperparameter tuning. 
Quantifying this difference with the performance score, Genetic Algorithm is improved by 0.195, PSO by 1.002, and Simulated Annealing by 0.366, for an overall average improvement of 204.7\% over the average limited hyperparameter configuration, and 210.8\% on the test set of search spaces.

\begin{figure}[tb]
    \centering
    \includegraphics[width=0.99\linewidth]{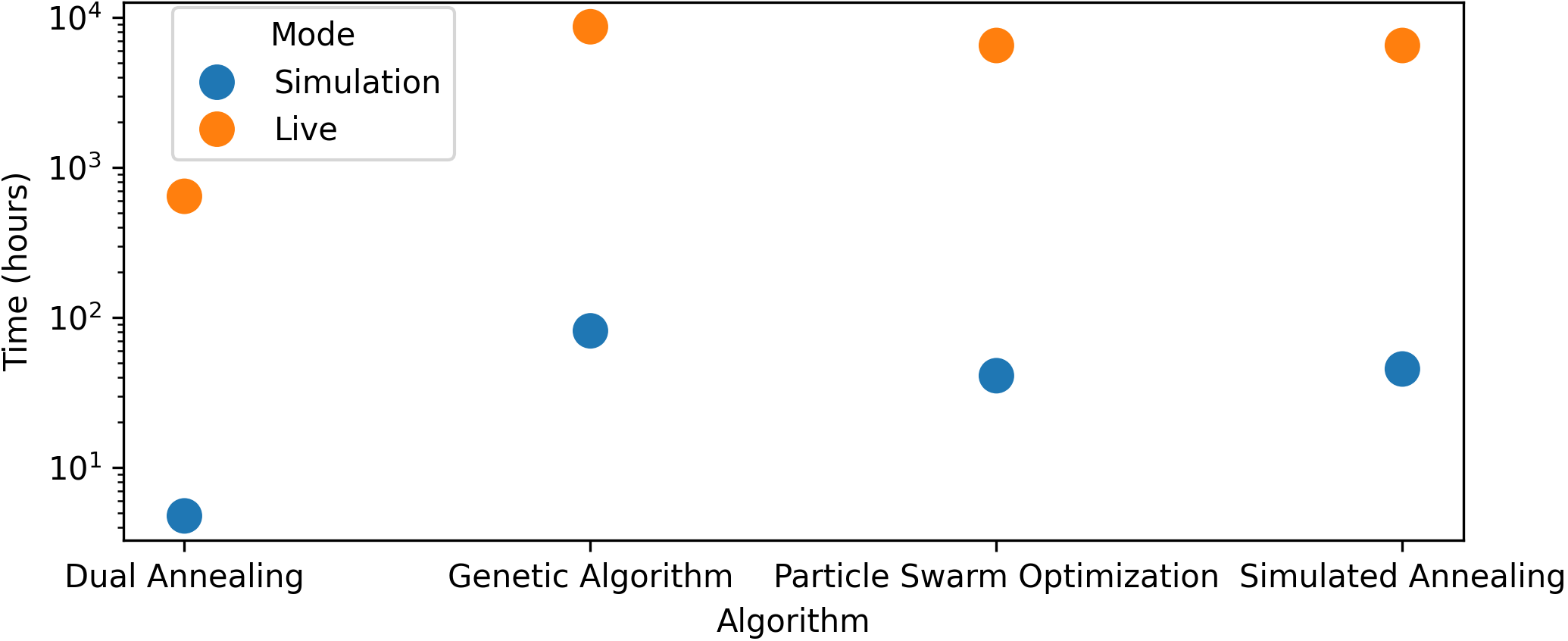}
    \vspace{-0.8cm}
    \caption{Tuning time comparison between live and simulation mode.} %
    \label{fig:results_tuning_time_comparison}
\end{figure}

\subsection{Feasibility with Simulation Mode}
\label{subsec:evaluation_results_scalability}
Finally, to evaluate the efficiency of our approach, we compare live tuning and using the simulation mode as shown in \cref{fig:results_tuning_time_comparison}, where lower is better. 
We calculate the time live tuning would have taken by taking the 95\% time budget of \cref{subsec:evaluation_setup} for each search space in seconds, multiplied by the number of hyperparameter configurations for each algorithm and the number of repeats (25 as per \cref{subsec:evaluation_setup}). 
With our method, exhaustive hyperparameter tuning of the limited set of \cref{tab:hyperparams} took approximately 4.5 hours for the optimization algorithm with the least hyperparameter configurations (\textit{Dual Annealing}) and 81 hours for the optimization algorithm with the most hyperparameter configurations (\textit{Genetic Algorithm}). %
While this requires the upfront cost of brute-forcing each search space, this is a one-off cost and allows the hyperparameter tuning of multiple optimization algorithms in parallel. 
In contrast, without the simulation mode, the exhaustive hyperparameter tuning would take approximately 642 hours (nearly a month) for the optimization algorithm with the least hyperparameter configurations and 8672 hours (nearly a year) for the optimization algorithm with the most hyperparameter configurations. 
In total, live-tuning the four optimization algorithms of this evaluation would take 22323 hours (nearly three years), whereas it took 172 sequential hours using the simulation mode, a 130x speedup. %

    \section{Conclusion} \label{sec:conclusion_futurework}

In this work, we have introduced a novel approach to hyperparameter tuning for optimization algorithms used in auto-tuning, a hitherto often overlooked domain. 
As processor architectures and applications continue to grow in complexity, efficient auto-tuning is crucial to fully leverage the computational power of modern architectures, which in turn depends on optimization algorithm performance. 
Our method addresses the challenges of hyperparameter optimization in this domain by building upon our statistically robust methodology for comparing optimization algorithms, introducing a simulation mode to make hyperparameter tuning feasible, and promoting the use of reproducible and shareable benchmark resources.

In this first evaluation of hyperparameter tuning for auto-tuning, the results show that our approach substantially improves the efficiency of the optimization algorithms in finding near-optimal configurations in auto-tuning search spaces and that these improvements hold on search spaces not trained on. 
By systematically tuning a limited set of hyperparameters for these algorithms with a robust performance score, we improved their overall performance by 94.8\% on average, and extensive tuning extended this to an average improvement of 204.7\%. 
Our simulation mode reduces the computational cost of hyperparameter tuning by two orders of magnitude, making large-scale experiments feasible and preventing the need for constant access to hardware and excessive resource usage. 
We show that the hyperparameters can be optimized efficiently using meta-strategies, allowing vast hyperparameter spaces to be tuned efficiently. 
The FAIR dataset of auto-tuning data provided in conjunction with this work lowers barriers to entry in auto-tuning research, as well as energy consumption and resource costs.
Based on the evaluation, we conclude that our method for hyperparameter tuning of auto-tuning optimization algorithms has a substantial effect on the performance, even on a limited set of hyperparameters, and generalizes well beyond the training data. 
By addressing the efficacy, efficiency, and reproducibility challenges in hyperparameter tuning, this work contributes to the advancement of auto-tuning, enabling more efficient utilization of modern processors in scientific and industrial applications.

While our approach achieves substantial performance improvements across multiple real-world applications and architectures, %
the reliance on exhaustively explored search spaces to ensure statistical robustness may limit applicability in cases where such data is unattainable. 
Future work will explore methods for extending this approach to partially explored or dynamically generated search spaces. %

The implementation and optimized hyperparameters in this work are included and set as defaults in Kernel Tuner, which can be installed with \verb|pip install kernel-tuner|. 
For more information, please visit the \href{https://github.com/KernelTuner/kernel_tuner}{repository}\footnote{\url{https://github.com/KernelTuner/kernel_tuner}}.

\clearpage

\bibliographystyle{IEEEtran}
\bibliography{references}

\end{document}